\documentclass[journal]{IEEEtran}
\IEEEoverridecommandlockouts
\usepackage{cite}
\usepackage{amsmath,amssymb,amsfonts}
\usepackage{algorithmic}
\usepackage{graphicx}
\usepackage{textcomp}
\usepackage{xcolor}
\usepackage{hyperref}
\usepackage{orcidlink}
\usepackage{multirow}
\usepackage{multicol}
\usepackage{graphicx}
\usepackage{booktabs}
\usepackage{booktabs}
\usepackage{array}
\usepackage{tabularx}
\def\BibTeX{{\rm B\kern-.05em{\sc i\kern-.025em b}\kern-.08em
    T\kern-.1667em\lower.7ex\hbox{E}\kern-.125emX}}
\usepackage{balance}

\begin{document}

\title{A Survey and Framework of Cooperative Perception: From Heterogeneous Singleton to Hierarchical Cooperation
}

\author{
Zhengwei~Bai$^{\orcidlink{0000-0002-4867-021X}}$,~\IEEEmembership{Student Member,~IEEE},
Guoyuan~Wu$^{\orcidlink{0000-0001-6707-6366}}$,~\IEEEmembership{Senior Member,~IEEE},
Matthew~J.~Barth$^{\orcidlink{0000-0002-4735-5859}}$,~\IEEEmembership{Fellow,~IEEE},
Yongkang~Liu,
Emrah Akin Sisbot,
Kentaro~Oguchi,
Zhitong Huang

\thanks{\textit{Corresponding Author}: Zhengwei Bai, E-mail: zbai012@ucr.edu.}

\thanks{Zhengwei Bai, Guoyuan Wu, and Matthew J. Barth are with the Department of Electrical and Computer Engineering, the University of California at Riverside, Riverside, CA 92507 USA .}

\thanks{Yongkang Liu, Emrah Akin Sisbot, and Kentaro Oguchi are with Toyota Motor North America, InfoTech Labs, Mountain View, CA 94043, USA.}

\thanks{Zhitong Huang is with Leidos Inc., McLean, VA, 22101, USA}
}

\maketitle

\begin{abstract}
Perceiving the environment is one of the most fundamental keys to enabling Cooperative Driving Automation (CDA), which is regarded as the revolutionary solution to addressing the safety, mobility, and sustainability issues of contemporary transportation systems. Although an unprecedented evolution is now happening in the area of computer vision for object perception, state-of-the-art perception methods are still struggling with sophisticated real-world traffic environments due to the inevitably physical occlusion and limited receptive field of single-vehicle systems. Based on multiple spatially separated perception nodes, Cooperative Perception (CP) is born to unlock the bottleneck of perception for driving automation. In this paper, we comprehensively review and analyze the research progress on CP and, to the best of our knowledge, this is the first time to propose a unified CP framework. Architectures and taxonomy of CP systems based on different types of sensors are reviewed to show a high-level description of the workflow and different structures for CP systems. Node structure, sensor modality, and fusion schemes are reviewed and analyzed with comprehensive literature to provide detailed explanations of specific methods. A Hierarchical CP framework is proposed, followed by a review of existing Datasets and Simulators to sketch an overall landscape of CP. Discussion highlights the current opportunities, open challenges, and anticipated future trends.
\end{abstract}


\section{Introduction}

The rapid progress of the transportation system has improved the efficiency of our daily people and goods movement. Nevertheless, the rapidly increasing number of vehicles has resulted in several major issues in the transportation system in terms of safety~\cite{2019Crash}, mobility~\cite{2018Congestion}, and environmental sustainability~\cite{2021Energy}. Taking advantage of recent strides in advanced sensing, wireless connectivity, and artificial intelligence, Cooperative Driving Automation (CDA) enables connected and automated vehicles (CAVs) to communicate between each other, with roadway infrastructure, or with other road users such as pedestrians and cyclists equipped with mobile devices, to improve the system-wide performance. Hence, CDA is attracting increasingly more attention over the past few years and is regarded as a transformative solution to the aforementioned challenges~\cite{fagnant2015preparing}. 


Object Perception (OP), acting as the ``vision'' function of automated agents by analogy, plays a fundamental role in the basic structure of CDA applications~\cite{2021SAE}. Different kinds of onboard or roadside sensors have different capabilities of perceiving the traffic conditions in the real-world environment. The perception data can act as the system input and support various kinds of downstream CDA applications, such as Collision Warning~\cite{wu2020improved}, Eco-Approach and Departure (EAD)~\cite{bai2022hybrid}, and Cooperative Adaptive Cruise Control (CACC)~\cite{wangCACC}.


With the development of sensing technologies, transportation systems can retrieve high-fidelity traffic data from different sensors. For instance, cameras can provide detailed vision data to classify various kinds of traffic objects, such as vehicles, pedestrians, and cyclists \cite{liu2020deep}. LiDAR can provide high-fidelity 3D point cloud data to grasp the precise 3D location of the traffic objects~\cite{arnold2019survey}. RADAR sensor has been an integral part of safety-critical applications in the automotive industry due to its robust performance in variable conditions~\cite{8443497}.

\begin{table*}[!h]
  \centering
  \caption{Relationship between classes of CDA cooperation and levels of automation~\cite{2021SAE}.}
  \resizebox{\textwidth}{!}{%
    \begin{tabular}{|c|p{5em}|p{7.5em}|p{7.5em}|p{7.5em}|p{7.5em}|c|c}
\cmidrule{3-8}    \multicolumn{1}{r}{} & \multicolumn{1}{c}{} & \multicolumn{6}{c}{\textbf{SAE Driving Automation (DA) Levels}} \\
\cmidrule{3-8}    \multicolumn{1}{r}{} & \multicolumn{1}{c|}{} & \textbf{Level 0:\newline  No DA} & \textbf{Level 1:\newline  Driver Assistance} & \textbf{Level2:\newline Partial DA} & \textbf{Level3:\newline  Conditional DA} & \multicolumn{1}{p{7.5em}|}{\textbf{Level 4:\newline  High DA}} & \multicolumn{1}{p{7.5em}}{\textbf{Level 5:\newline  Full DA}} \\
\cmidrule{2-8}     \parbox[t]{2mm}{\multirow{5}{*}{\rotatebox[origin=c]{90}{\textbf{CDA Cooperation Classes\space\space\space\space\space\space\space}}}} & \textbf{No Cooperation} & e.g., Signage & \multicolumn{2}{p{18em}|}{Relies on driver to supervise performance in real-time} & \multicolumn{3}{p{22.5em}}{Relies on ADS under defined conditions} \\
\cmidrule{2-8}          & \textbf{Class A:\newline states-sharing} & e.g., Traffic Signal & \multicolumn{2}{p{18em}|}{Limited Cooperation: Human is driving and supervise CDA features} & \multicolumn{3}{p{22.5em}}{Improved C-ADS situational awareness by on-board sensing and surrounding roadusers and operators} \\
\cmidrule{2-8}          & \textbf{Class B:\newline intent-sharing} & e.g., Turn Signal & Limited Cooperation (only longitudinal OR alteral) & Limited Cooperation (both longitudinal AND alteral) & \multicolumn{3}{p{22.5em}}{Improved C-ADS situational awareness through prediction reliability} \\
\cmidrule{2-8}          & \textbf{Class C:\newline agreement-sharing} & e.g., Hand Signals & N/A   & N/A   & \multicolumn{3}{p{22.5em}}{Improved Ability of C-ADS by coordination with surrounding road users and operators} \\
\cmidrule{2-8}          & \textbf{Class D:\newline prescriptive} & e.g., Lane Assignment & N/A   & N/A   & \multicolumn{3}{p{22.5em}}{C-ADS has full authority to decide actions except for very specific cases} \\
\cmidrule{2-8}    \end{tabular}}%
  \label{tab:cda}%
\end{table*}%

During the last couple of decades, a large portion of the OP methods and high-fidelity perception data have come from onboard sensors while most of the roadside sensors are still used for traditional traffic data collection such as counting traffic volumes based on loop detectors, cameras, or radars~\cite{zou2019object}. Although empowered with advanced perception methods, onboard sensors are inevitably limited by the range and occlusion by other objects. Infrastructure-based perception systems have the potential to achieve better OP results with fewer occlusion effects and more flexibility in terms of mounting height and pose. However, due to the fixture of installation, infrastructure-based sensors will suffer from limited receptive ranges and sometimes large blind zones. Thus, neither onboard sensors nor infrastructure-based sensors alone can outbreak the physical limitations and achieve satisfactory perception performance.

Empowered by mobile connectivity, Connected Vehicles (CVs) and Connected and Automated Vehicles (CAVs) have the capability to grasp perception information from others who are equipped with perception systems and connectivity, such as smart infrastructures or other CAVs. It is reasonable to combine sensing information from spatially separated nodes to overcome the occlusion or perception range. Thus, Cooperative Perception naturally attracts fast-increasing attention to Driving Automation. Many kinds of research have been conducted from different aspects, such as perception nodes (vehicle~\cite{F-cooper} or infrastructure~\cite{bai2022pillargrid}), sensor modalities (Camera~\cite{zhu2021mme} or Lidar~\cite{bai2022cmm}), and fusion schemes (early fusion~\cite{chen2019cooper}, late fusion~\cite{arnold2020cooperative}, or deep fusion~\cite{xu2022v2x}). Although a recent survey conducted by Caillot et~al.~\cite{caillot2022survey} reviewed the cooperative perception in an automotive context, their focus is mainly on the ego-vehicle, such as localization, map generation, etc. Thus, a comprehensive overview of CP and a general CP framework for handling heterogeneity and scalability in mixed traffic are still missing.

In this paper, the CP-based object perception methods are reviewed, which aims to establish an overall landscape for cooperative perception based on different aspects including 1) node structures, 2) sensor modalities, and 3) fusion schemes. Furthermore, a hierarchical CP framework is proposed to unify different scenarios in terms of different perspectives mentioned above and to provide inspiration for future research. 

The rest of this paper is organized as follows: Architectures and taxonomy for CP systems are reviewed in Section~\ref{sec: architecture} to lay the foundation. Major pillars including node structure, sensor modality, and fusion scheme are reviewed in Section~\ref{sec: Node}~to~\ref{fusion}, respectively, followed by the presentation of available Datasets and Simulators. The hierarchical cooperative perception framework is proposed and discussed in Section~\ref{sec: hcp}. Section~\ref{sec: dis} highlights the current challenges and future trends, followed by Section~\ref{sec: conc} that concludes the paper.

\section{Architecture and Taxonomy}
\label{sec: architecture}

For the development of driving automation, the Society of Automotive Engineers (SAE) initiated the SAE J3016 Standard, commonly known as the \textit {SAE Levels of Driving Automation}~\cite{J3016_202104}, which has been the fundamental source guiding the development of driving automation. Six levels of driving automation are classified from Level 0 (No driving automation) to Level 5 (Full driving automation) in terms of motor vehicles. Defined by the SAE J3216 Standard~\cite{2021SAE}, Cooperative Driving Automation (CDA) enables communication and cooperation between equipped vehicles, infrastructure, and other road users, which will, in turn, improve the safety, mobility, and sustainability of transportation systems. By further extending the SAE levels of Driving Automation, SAE J3216 defines the CDA levels into five classes including 1) No cooperative automation, 2) Class A: Status-sharing, 3) Class B: Intent-sharing, 4) Class C: Agreement-seeking, and 5) Class D: Prescriptive. Table\ref{tab:cda} summarized the details and relationship between classes of CDA cooperation and levels of driving automation. According to Table~\ref{tab:cda}, cooperative perception plays a significant and fundamental role in supporting both CDA and Automated Driving systems. Led by the Federal Highway Administration, the CARMA program ~\cite{2021CARMA} is one of the state-of-the-art (SOTA) projects that aim to support and enable research and testing for CDA. Based on the analysis of SAE standards and the CARMA program, the architecture and taxonomy of CP are proposed and described in the following sections.

\subsection{Architecture}

\begin{figure*}[!ht]
    \centering
    \includegraphics[width=0.95\textwidth]{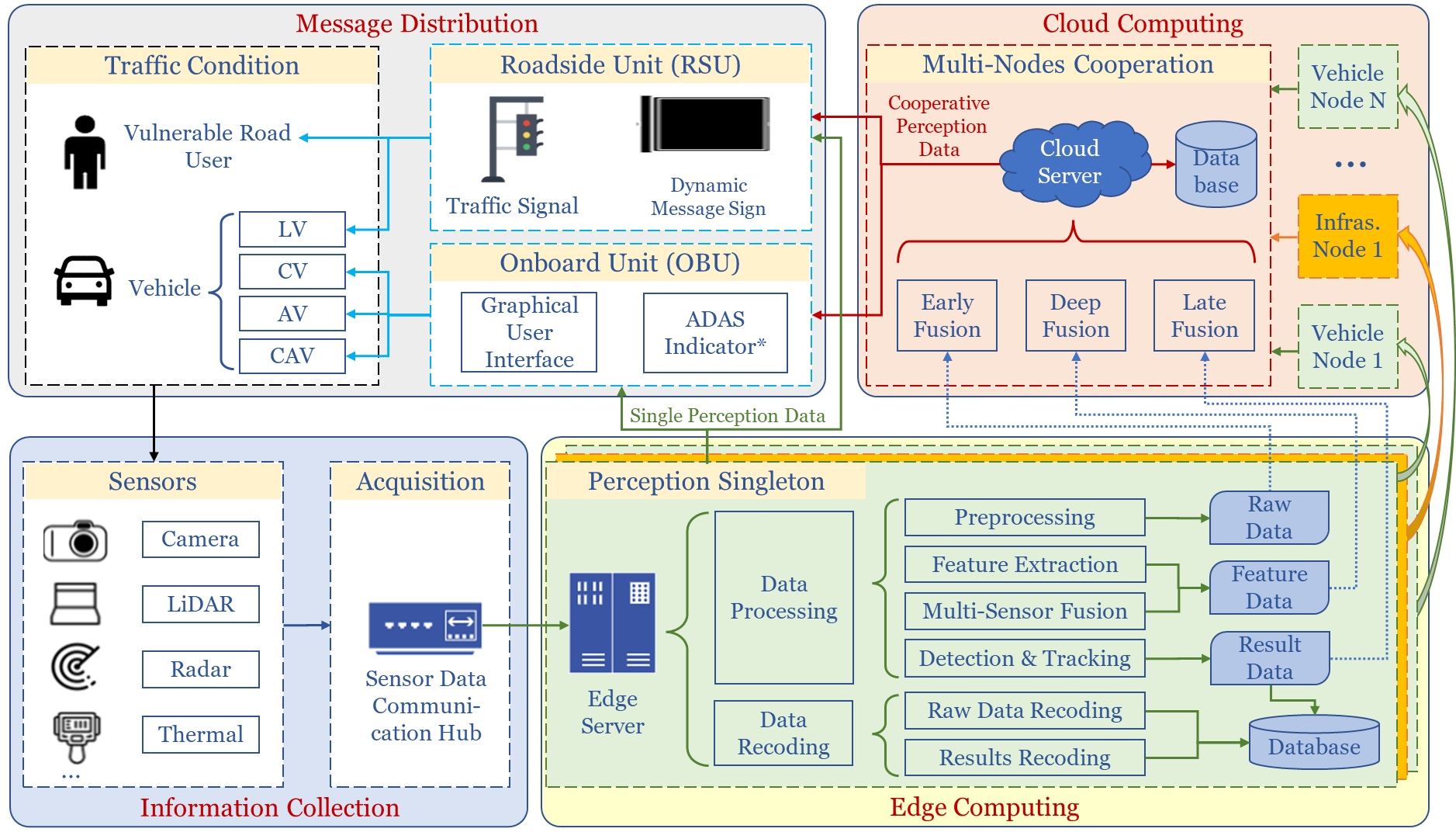}
    \caption{Systematic architecture for cooperative perception system (*: Other non-visual driving advisory signals for Advanced driver-assistance systems (ADAS), such as audial, haptic, or even control commands.).}
    \label{fig: roadside perception structure}
\end{figure*}

In cooperative driving automation (CDA), the fidelity and range of perception information have a significant impact on the system performance of subsequent cooperative maneuvers. Fig.~\ref{fig: roadside perception structure} demonstrates a system architecture of the cooperative perception system for enabling CDA. Specifically, four typical phases can be identified in the CP process: 1) \textit{Information Collection}; 2) \textit{Edge Processing}; 3) \textit{Cloud Computing}; and 4) \textit{Message Distribution}.
    
\subsubsection{Information Collection}
Collecting raw data of traffic information lays the foundation for downstream perception tasks. In the development of transportation, various kinds of sensors are implemented aiming at different tasks and scenarios. In the context of traffic surveillance, several traditional sensors are widely applied, such as \textit{Loop Detectors} and \textit{Microwave RADAR}, for dynamic traffic management~\cite{nellore2016survey}. However, the main capacity of these traditional sensors is to provide mesoscopic traffic information, such as traffic volume or queue length. To support cooperative driving automation, 3D object-level information is required, which can be generated from high-resolution sensors, such as cameras, LiDAR, etc. 

Dated back to a couple of decades ago, due to the limitation of computational power and development of the computer vision field, high-resolution-sensors-based object perception is barely developed for intelligent transportation systems ~\cite{cheung2005traffic}. Although some vision-based methods are developed, very limited performance can be achieved~\cite{COIFMAN1998271}. Thanks to the quick advancement in high-performance computation and the surge of artificial intelligence (AI)~\cite{lecun2015deep},  high-resolution sensors are able to provide object-level perception results, which can be equipped on 
vehicles or roadside infrastructures to perceive the environment and transmit collected data to its processing server via a communication hub for further processing.

\subsubsection{Edge Processing}
\label{edge}
Unlike traditional traffic surveillance systems which do not need high-frequency and low-latency processing, CDA generally requires perception data with a minimal $1-10Hz$ frequency and time delay of less than $100ms$~\cite{v2x8466351}. Considering that using limited bandwidth to transmit a large volume of raw data (e.g., point cloud data) may cause an unacceptable time delay (especially in some safety-critical scenarios), information collected from sensors may be processed on edge servers equipped on vehicles or infrastructures. In this paper, the singleton empowered with perception and communication capabilities is regarded as a \textit{Perception Node} (PN). Generally, there are six main steps for processing the raw sensing data at a single PN~\cite{bai2022cmm}, as shown below:
\begin{itemize}
    \item \textit{Preprocessing}: Manipulations of raw data to provide a ready-to-use format for perception modules with respect to specific sensors, such as coordinate transformation, geo-fencing, and noise reduction.
    \item \textit{Feature Extraction}: Feature extraction for subsequent perception task by applying deep neural networks (DNNs) or traditional statistical methods.
    \item \textit{Multi-Sensor Fusion}: Multi-sensor fusion algorithms may be applied if there is more than one sensor used for a single PN.
    \item \textit{Detection \& Tracking}: Generation of object detection and tracking results for demonstrating position, pose, and identification of certain road users, such as rotated bounding boxes with unique IDs and classification tags. 
    \item \textit{Raw Data Logging}: Recording of raw sensing data with timestamps for post-analysis.
    \item \textit{Results Logging}: Recording of semantic perception data with timestamps for post-analysis.
\end{itemize}

Different types of PNs play different roles in a CP system. For a Vehicle PN (V-PN), edge computing mainly serves itself, i.e., perceiving the environment to support the downstream driving tasks such as decision-making or control. For an Infrastructure PN (I-PN), its main purpose is to improve the situation awareness at a fixed location by advanced ranging sensing (e.g., camera, LiDAR) and communications.
 
\subsubsection{Cloud Computing}
\label{cloud computing}
Considering the large-scale implementation of cooperation, cloud computing is involved to act as the fusion center for multiple PNs. Information from heterogeneous PNs will be transmitted to the \textit{Cloud} via different kinds of communications. For mobile road users (e.g., vehicles, cyclists, pedestrians), wireless communication, such as \textit{Cellular Network}, \textit{Wireless Local Area Network (WLAN)}, etc. is used to exchange information with the \textit{Cloud}. Additionally, infrastructure can take advantage of both wireless and wired communications (e.g., \textit{Optical Fiber}, \textit{Local Area Network (LAN)}, etc) by well balancing the cost and system performance such as delay~\cite{dey2016vehicle}.

Generally, three types of perception data are generated from heterogeneous PNs: 
\begin{itemize}
    \item Raw data which contains the original information from sensors, e.g., RGB images from the camera, point cloud data (PCD) from LiDAR, etc.
    \item Feature data which contains the hidden feature extracted by neural network or statistical methods for representing the raw data in higher dimensional spaces.
    \item Result data which contains the semantic perception information such as 2D/3D location, size, rotation, etc.
\end{itemize}

One of the key components for CP is data fusion and different fusion schemes will be applied, depending on the types of data to be shared between PNs and the \textit{Cloud}. For instance, early fusion, deep fusion, and late fusion are based on raw data, feature data, and result data, respectively. Due to the limited bandwidth of wireless communication, result data are most widely used for CP or other CDA tasks~\cite{bai2022cmm}. A few systems that have high-speed communication capability, which allow high-volume low-latency data transmission, can also transmit raw data to the Cloud for processing, and some work has been conducted to enhance driving automation~\cite{chen2019cooper}. In terms of multi-node perception systems, i.e., simultaneously perceiving the environment from different locations, time alignment (with the necessity of delay compensation) and object association need to be considered for spatiotemporal information assimilation and synchronization. Recently, deep fusion (also named intermediate fusion) attracts increasingly popular attention due to its superiority in CP performance~\cite{bai2022pillargrid, xu2022v2x}. Detailed illustration and literature review for fusion schemes are conducted in Section~\ref{fusion}.


\subsubsection{Message Distribution}

The perception information (along with advisory or actuation signals) can be distributed to road users in two major ways, depending on the connectivity status. For conventional road users without wireless connectivity, such information can be delivered to end devices at the roadside, such as \textit{Dynamic Message Sign} (DMS) or signal head display of traffic lights via the Traffic Management Center (TMC). For road users with connectivity, customized information, e.g., surrounding objects and \textit{Signal Phase and Timing} (SPaT) of upcoming signals, and various visual/non-visual ADAS indicators can be accessed to enable various connected driving automation applications, such as Connected Eco-Driving~\cite{altan2017glidepath, bai2022hybrid}. Cooperative Perception messages can support more sophisticated cooperative maneuvers in a mixed traffic environment. For example, vulnerable road users (VRUs) and legacy vehicles (LVs) can react to the message shown on DMS~\cite{wu2020improved}. Connected vehicles (CVs) can use CP information to get better situational awareness and pass through intersections in a safer manner~\cite{bai2022cyber}. Autonomous vehicles (AVs) and connected and automated vehicles (CAVs) can improve their driving performance via better coordination algorithms~\cite{shan2020demonstrations}.

\subsection{Taxonomy}
\begin{figure}[!ht]
    \centering
    \includegraphics[width=0.45\textwidth]{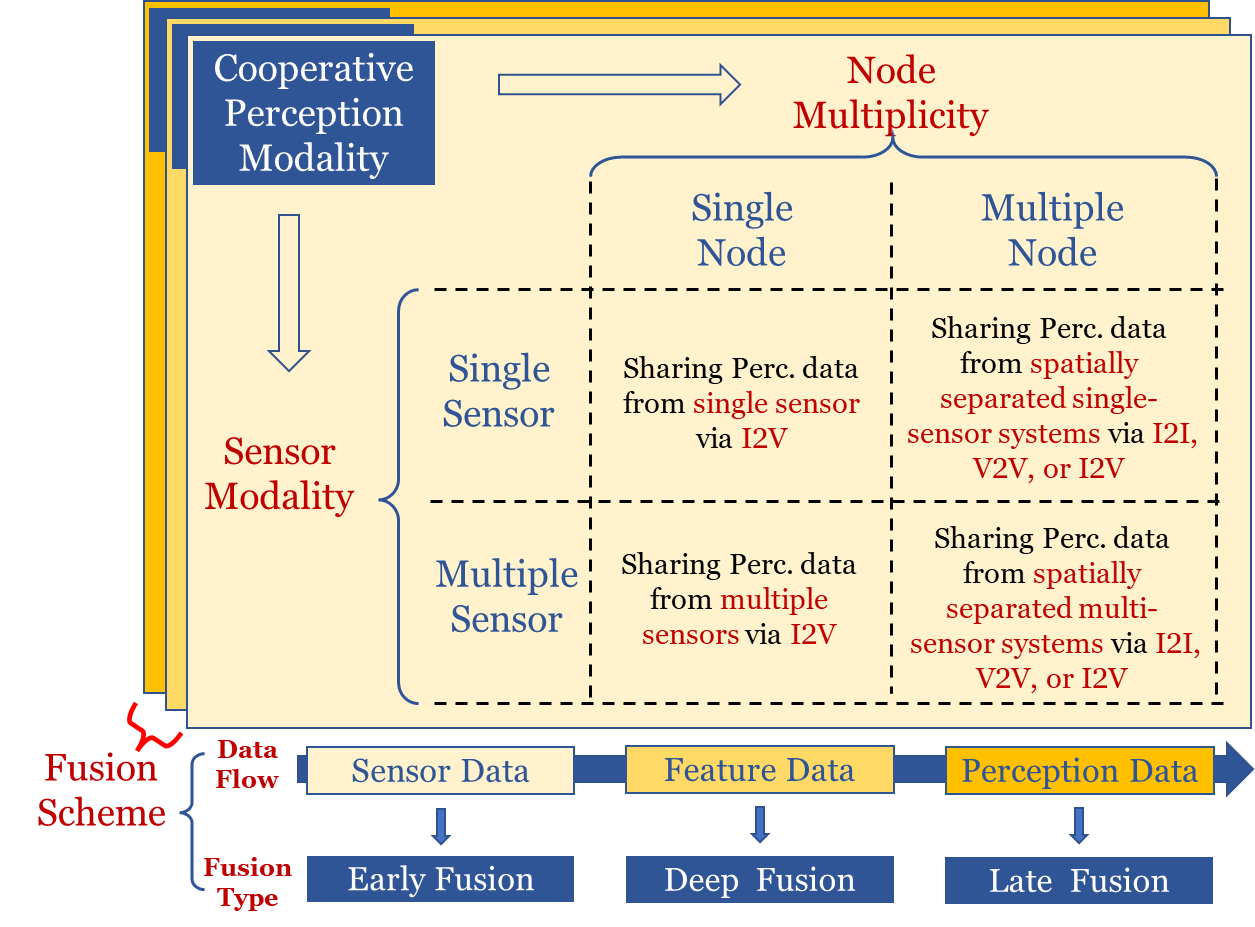}
    \caption{Taxonomy of CP in terms of node multiplicity, sensor modality, and fusion scheme.}
    \label{fig: taxonomy}
\end{figure}
Based on the architecture of CP illustrated above, three key aspects are identified for a CP system, including 1) Node Multiplicity, 2) Sensor Modality, and 3) Fusion Scheme, and Fig.~\ref{fig: taxonomy} illustrates these aspects in detail. In terms of node multiplicity and sensor modality, four types of CP systems can be identified as follows:
\begin{itemize}
    \item \textit{Single-Node Single-Mode CP (SS-CP)}: Cooperation between heterogeneous PNs by sharing perception data from the single-modal sensor(s) via infrastructure-to-everything (I2X) or vehicle-to-everything (V2X) communications.
    \item \textit{Multi-Node Single-Mode CP (MS-CP)}: Cooperation between heterogeneous PNs by sharing perception data from single-modal multiple sensors perception via I2X and/or V2X communications.
    \item \textit{Single-Node Multi-Mode CP (SM-CP)}: Cooperation between heterogeneous PNs by sharing perception data from multi-modal sensor perception via I2X or V2X communications.
    \item \textit{Multi-Node Multi-Mode CP (MM-CP)}: Cooperation between heterogeneous PNs by sharing perception data from multi-modal sensor perception via I2X and/or V2X communications.
\end{itemize}
For each of the four CP types, three fusion schemes can be applied based on the types of perception data, which have been introduced in Section~\ref{cloud computing}. In the following, a comprehensive literature review is conducted with detailed analyses on the aspects of node multiplicity, sensor modality, and fusion scheme, respectively.  

\section{Node Structure}
\label{sec: Node}
In this paper, we define \textit{Node} to be a Perception Node (PN) that is capable of perceiving and communicating -- as the fundamental unit for building the CP system. As mentioned in Section~\ref{fig: taxonomy}, CP systems can be divided into Single-Node and Multi-Node CP systems. Meanwhile, in Section~\ref{edge}, the vehicle node (V-PN), and the infrastructure node (I-PN) are considered heterogeneous nodes in CP systems. For comprehensiveness and conciseness, CP is discussed from the aspect of \textit{Node Structure} in this section: 1) I-PN-based CP, 2) V-PN-based CP, and 3) heterogeneous-PN-based CP.

\subsection{I-PN-based CP}

Object perception based on roadside sensors has a great potential to break the current bottleneck for autonomous driving, especially in a mixed traffic environment via cooperative perception~\cite{gupta2021deep}. This section reviews the infrastructure-based object detection and tracking approaches in the literature. 
\subsubsection{Camera-based I-PN}
Infrastructure-based camera systems have been widely used for object detection and a survey conducted by Zou~et~al.~\cite{zou2019object} shows various camera-based applications in traffic scenes, such as traffic surveillance, safety warning, traffic management, etc. 
Monovision camera plays a significant role in object detection. Ojala~et~al. proposed a \textit{Convolutional Neural Network} (CNN) based pedestrian detection and localization approach using roadside cameras ~\cite{Ojala8793228}. The perception system consists of a monovision camera streaming video and a computing unit that performs object detection and positioning. Besides, Guo~et~al. proposed a 3D vehicle detection method based on a monocular camera~\cite{9502706}, which consists of three steps: 1) clustering arbitrary object contours into linear equations; 2) estimating positions, orientations, and dimensions of vehicles by applying the K-means method; and 3) refining 3D detection results by maximizing a posterior probability. 

Instead of using a fixed roadside camera, some researchers try to take advantage of Unmanned Aerial Vehicle (UAV) based cameras.  MultEYE~\cite{balamuralidhar2021multeye} is a monitoring system for real-time vehicle detection, tracking, and speed estimation proposed by Balamuralidhar~et~al. Different from general roadside sensors equipped on signal poles or light poles, the data source of MultEYE comes from an Unmanned Aerial Vehicle (UAV) equipped with an embedded computer and a video camera. Inspired by the multi-task learning methodology, a segmentation head~\cite{paszke2016enet} is added to the object detector backbone~\cite{bochkovskiy2020yolov4}. Dedicated object tracking~\cite{bolme2010visual} and speed estimation algorithms have been optimized to track objects reliably from a UAV with limited computational efforts. Cicek and G{\"o}ren proposed a deep-learning-based automated curbside parking spot detection approach through a roadside camera~\cite{cicek2021fully}. To identify the road boundaries, object detection and road segmentation methods are employed by utilizing \textit{the FCN-VGG16} model~\cite{long2015fully} on the \textit{KITTI} dataset~\cite{Geiger2012CVPR} and \textit{Faster R-CNN}~\cite{ren2015faster} on \textit{MS-COCO} dataset~\cite{lin2014microsoft}, respectively. Then, a method is designed to differentiate parked vehicles from the moving ones and then give them guidance on the nearest spot information to drivers.

For multi-camera perception systems from the roadside, Arnold~et~al. proposed a cooperative 3D object detection model by utilizing multiple depth cameras to mitigate the limitation of field-of-view (FOV) of a single-sensor system~\cite{arnold2020cooperative}. For each camera, a depth image is projected to pseudo-point-cloud data~\cite{ glennie2010static}. Two sensor-fusion schemes are designed: early fusion and late fusion (see Fig.~\ref{fig: pipelines}) and adapted based on Voxelnet~\cite{zhou2018voxelnet}. The evaluation in a T-junction and a roundabout scenario in the CARLA simulator~\cite{dosovitskiy2017carla} demonstrates that the proposed method can enlarge the detection coverage without compromising accuracy.

\subsubsection{LiDAR-based I-PN}
In recent years, roadside LiDAR sensors attract increasing attention from researchers about object perception in transportation. Using roadside LiDAR, Zhao~et~al. proposed a detection and tracking approach for pedestrians and vehicles~\cite{zhao2019detection}. As one of the early studies utilizing roadside LiDAR for perception, a classical detection and tracking pipeline for PCD was designed. It mainly consists of 1) \textit{Background Filtering}: To remove the laser points reflected from road surfaces or buildings by applying a statistics-based background filtering method~\cite{wu2017automatic}; 2) \textit{Clustering}: To generate clusters for the laser points by implementing a DBSCAN method~\cite{ester1996density}; 3) \textit{Classification}: To generate different labels for different traffic objects, such as vehicles and pedestrians, based on neural networks~\cite{li2012brief}; and 4) \textit{Tracking}: To identify the same object in continuous data frames by applying a discrete Kalman filter~\cite{welch1995introduction}. Based on the aforementioned work, Cui~et~al. designed an automatic vehicle tracking system by considering vehicle detection and lane identification~\cite{cui2019automatic}. A real-world operational system is developed, which consists of a roadside LiDAR, an edge computer, a \textit{Dedicated Short-Range Communication} (\textit{DSRC}) \textit{Roadside Unit} (\textit{RSU}), a Wi-Fi router, and a DSRC \textit{On-board Unit} (\textit{OBU}), and a GUI. Following a similar workflow, Zhang~et~al. proposed a vehicle tracking and speed estimation approach based on a roadside LiDAR~\cite{zhang2020vehicle}. Vehicle detection results are generated by the ``\textit{Background Filtering-Clustering-Classification}'' process. Then, a centroid-based tracking flow is implemented to obtain initial vehicle transformations, and the unscented Kalman Filter~\cite{julier2004unscented} and joint probabilistic data association filter~\cite{bar2009probabilistic} are adopted in the tracking flow. Finally, vehicle tracking is refined through a Brid-Eye-View (BEV) LiDAR-image matching process to improve the accuracy of estimated vehicle speeds. Following the bottom-up pipeline mentioned above, numerous roadside LiDAR-based methods are proposed from various points of view~\cite{zhang2020gc, Song9216093, gouda2021automated, 8484040, zhang2019automatic}.

On the other hand, using learning-based models to cope with LiDAR data is another main methodology. Bai~et~al.~\cite{bai2022cyber} proposed a deep-learning-based real-time vehicle detection and reconstruction system from roadside LiDAR data. Specifically, CARLA simulator~\cite{dosovitskiy2017carla} is implemented for collecting the training dataset, and ComplexYOLO model~\cite{simony2018complex} is applied and retrained for the object detection on the CARLA dataset. Finally, a co-simulation platform is designed and developed to provide vehicle detection and object-level reconstruction, which aims to empower subsequent CDA applications with readily retrieved authentic detection data. In their following work for real-world implementation, Bai~et~al.~\cite{bai2022cmm} proposed a deep-learning-based 3D object detection, tracking, and reconstruction system for real-world implementation. The field operational system consists of three main parts: 1) 3D object detection by adopting PointPillar~\cite{lang2019pointpillars} for inference from roadside PCD; 2) 3D multi-object tracking by improving DeepSORT~\cite{veeramani2018deepsort} to support 3D tracking, and 3) 3D reconstruction by geodetic transformation and real-time onboard \textit{Graphic User Interface} (\textit{GUI}) display. 

By combining traditional and deep learning algorithms Gong~et~al.~\cite{gong2021pedestrian} proposed a roadside LiDAR-based real-time detection approach. Several techniques are designed to guarantee real-time performance, including the application of Octree with region-of-interest (ROI) selection, and the development of an improved Euclidean clustering algorithm with an adaptive search radius. The roadside system is equipped with \textit{NVIDIA Jetson AGX Xavier}, achieving the inference time of 110 ms per frame.




\begin{table*}[!ht]
  \centering
  \caption{Summary of Different Node Structures for Cooperative Perception.}
  \resizebox{\textwidth}{!}{%
    \begin{tabular}{c|c|p{20em}|c|c}
    \toprule
    \multicolumn{1}{c|}{Structure} & 
    \multicolumn{1}{c|}{Modality} & 
    \multicolumn{1}{c|}{Pros. and Cons.} & \multicolumn{1}{c|}{Highlighted Features} & \multicolumn{1}{c}{Author} \\
    \midrule
    \multicolumn{1}{c|}{\multirow{4}[8]{*}{Single-Node}} & \multicolumn{1}{c|}{\multirow{2}[4]{*}{Infrastructure}} & Pros: Higher location with flexible pose leads to less occlusion and system-level cost-effective.  & \multicolumn{1}{c|}{\multirow{2}[4]{18em}{Infrastructure assisted high-fidelity traffic surveillance}} & \multicolumn{1}{c}{\multirow{2}[4]{*}{Bai~et~al.~\cite{bai2022cmm}}} \\
\cmidrule{3-3}      &   & Cons: Need infrastructure support. &   &  \\
\cmidrule{2-5}      & \multicolumn{1}{c|}{\multirow{2}[4]{*}{Vehicle}} & Pros: Low latency perception for ego-vehicle. & \multicolumn{1}{c|}{\multirow{2}[4]{18em}{Everything on the vehicle side: sensing, processing, analysis.}} & \multicolumn{1}{c}{\multirow{2}[4]{*}{Arnol~et~al.~\cite{arnold2019survey}}} \\
\cmidrule{3-3}      &   & Cons: Easily occluded by the surrounding vehicles or buildings. &   &  \\
    \midrule
    \multicolumn{1}{c|}{\multirow{6}[12]{*}{Multi-Node}} & \multicolumn{1}{c|}{\multirow{2}[4]{*}{Vehi. + Vehi.}} & Pros: Extend perception range from vehicle side. & \multicolumn{1}{c|}{\multirow{2}[4]{18em}{Sharing features generated from convolutional neural networks.}} & \multicolumn{1}{c}{\multirow{2}[4]{*}{Chen~et~al.~\cite{F-cooper}}} \\
\cmidrule{3-3}      &   & Cons: Occlusion by other vehicles. &   &  \\
\cmidrule{2-5}      & \multicolumn{1}{c|}{\multirow{2}[4]{*}{Infra. + Infra.}} & Pros: Extend perception range from infrastructure side. & \multicolumn{1}{c|}{\multirow{2}[4]{18em}{Sharing preprocessed RGB data among all roadside sensors.}} & \multicolumn{1}{c}{\multirow{2}[4]{*}{Arnold~et~al.~\cite{arnold2020cooperative}}} \\
\cmidrule{3-3}      &   & Cons: Have blind zone under the sensor. &   &  \\
\cmidrule{2-5}      & \multicolumn{1}{c|}{\multirow{2}[4]{*}{Infra. + Vehi.}} & Pros: Achieve a comprehensive range and field of view (FOV) for perception. & \multicolumn{1}{c|}{\multirow{2}[4]{18em}{Considering asynchronous
information sharing, pose errors, and heterogeneity of V2X components.
}} & \multicolumn{1}{c}{\multirow{2}[4]{*}{Xu,~et~al.~\cite{xu2022v2x}}} \\
\cmidrule{3-3}      &   & Cons: Require heterogeneity of the model. &   &  \\
    \bottomrule
    \end{tabular}}%
  \label{tab: node}%
\end{table*}%

\subsection{Vehicle Nodes}

Cooperative perception between vehicles mainly emerged from the research for Unmanned Aerial Vehicles (UAVs) to provide estimated localization in the region of interest. Back in 2006, Merino~et~al.~\cite{merino2006cooperative} proposed a multi-UAV CP system based on a distributed-centralized CP framework (similar to the current ``edge-cloud'' framework). The sensor data (such as images) collected from UAVs will be processed on the UAV side including image segmentation, stabilization of sequences of images, and geo-referencing. The location of objects in the region of interest will be estimated by UAVs and then send to a central server for further fusion by utilizing a probabilistic model.

For on-road vehicles, Rockl~et~al.~\cite{rockl2008v2v} propose a \textit{Multi-Sensor Multi-Target Tracking} method by associating the received sensor data via V2V communication. A more notable CP system for on-road vehicles was proposed by Rauch~et~al.~\cite{Car2X6232130} in 2012. A Car2X-based module was proposed to cooperate the perception results jointly for both spatial and temporal dimensions via the Unscented Kalman filter (UKF). Specifically, the object data shared from other vehicles need to be aligned to the coordinate of the host vehicle and synchronized in time.  Machine~et~al.~\cite{Machine8569832} proposed a machine learning-based method to fuse proposals generated by different connected agents. A specific center-point estimation method was proposed for generating the object location into the coordinate system of the host vehicle. Xiao~et~al.~\cite{xiao2018multimedia} proposed a CP method by sharing semantic segmentation information generated by a DNN and vision-feature matching data from the BEV-projected image data. GPS data was required for spatial alignment.

A comprehensive autonomous driving system (ADS) was implemented by Kim~et~al.~\cite{kim2014multivehicle}, whose core innovation is a CP system that provides ego-vehicle information beyond occlusion by a leading vehicle. A real-world system was deployed to validate the effectiveness of CP for enabling driving automation in multiple tasks, such as the forward collision warning, overtaking/lane-changing assistance, automated lane-change capability, etc. Experiments demonstrated that by enabling ego-vehicle with expanded perception information, the potential of driving automation can be significantly improved.

For CP system based on LiDAR data, Chen~et~al. proposed an early fusion method (\textit{Cooper}~\cite{chen2019cooper}) by aligning raw point cloud data (PCD) from multiple vehicles. To fulfill the limited bandwidth of V2V communication, raw PCD was preprocessed to reduce its size. Additionally, GPS and \textit{Inertial Measurement Unit} (IMU) data were required for PCD alignment. Then a PCD detector was designed based on VoxelNet~\cite{zhou2018voxelnet}, Sparse Convolution~\cite{yan2018second}, and Region Proposal Network (RPN)~\cite{ren2016faster}. The experiments demonstrated that Cooper was capable of improving perception performance by expanding sensing data. Following the \textit{Cooper}, Chen~et~al. proposed \textit{F-Cooper}~\cite{F-cooper}, a feature-based CP system using PCD. The core idea of F-Cooper is a two-step process: 1) to extract the hidden feature from sensor data via a DNN at each vehicle side, i.e., V-PN; 2) to generate perception results based on cross-vehicle feature data sharing.

CNN-based feature sharing was also applied in the work proposed by Marvasti~et~al.~\cite{marvasti2020cooperative} for the V2V CP task, named \textit{Feature Sharing Cooperative Object Detection} (FS-COD). Both FS-COD and F-Cooper complete spatial alignment at the feature level. However, different from F-Cooper which uses \textit{maxout} operation~\cite{goodfellow2013maxout} (i.e., output maximum value for corresponding multi-source data points) to fuse the multi-source data, FS-COD uses summation for multi-source feature fusion.

Considering compressing the feature data for transmission, Wang~et~al. proposed V2VNet~\cite{wang2020v2vnet}, which leverages the power of both deep neural networks and data compression. Specifically, a pipeline of  ``\textit{feature extraction-compression-decompression-object detector}'' is created to further consider the limitation of communication. Additionally, a novel simulator, \textit{Lidarsim}~\cite{manivasagam2020lidarsim}, is involved for cooperative perception to generate a PCD-based V2V dataset in a more realistic manner.

Zhang~et~al.~\cite{zhang2021distributed} proposed a  vehicle-edge-cloud framework for dynamic map fusion. Federated learning is applied for generating object detection results from multiple V-PNs and a three-stage fusion scheme is proposed to generate the final objects based on overlapping results from multiple PNs.

Xu~et~al.~\cite{xu2021opv2v} propose a feature-sharing-based CP model by V2V communication. Vehicles' relative pose information with respect to ego-vehicle is required for spatial alignment and feature generation. Specifically, the attention operation~\cite{vaswani2017attention} is applied for multi-node feature fusion and an open-source simulation-based dataset is developed and implemented for model training and validation.

\subsection{Heterogeneous PN-based CP}
Although many researchers have dug into cooperative perception from the perspectives of infrastructure perception and V2V cooperation, so far, only a few pieces of research are conducted for CP between heterogeneous PNs, i.e., cooperation between vehicles and infrastructure.

For cooperation between vehicles and infrastructure, Bai~et~al.~\cite{bai2022pillargrid} proposed a CP method, named  \textit{PillarGrid}, to generate 3D object detection results by PCD from onboard-roadside LiDAR sensors. Specifically, decoupled multi-stream CNNs are applied for feature extraction. The vehicle pose information is required for spatial alignment and the feature data are shared via V2X communication. A Grid-wise Feature Fusion (GFF) method is proposed for multi-PN feature fusion, which endows the PillarGrid with better scalability and capacity to handle heterogeneity.

Using Vision Transformer (ViT)~\cite{dosovitskiy2020vit}, Xu~et~al.~\cite{xu2022v2x} proposed a CP method named \textit{V2X-ViT}, which applied a share-weights CNNs for feature extraction. Ego-vehicle pose information is transmitted to surrounding vehicles and infrastructures for raw data alignment. Heterogeneous Graph Transformer (HGT)~\cite{hu2020heterogeneous} is designed to deal with different feature fusion types, e.g., V2V, V2I, etc. A window attention module is designed to capture hidden features from the fused feature map, which is then used to generate the object detection results.

\subsection{Summary}
Table~\ref{tab: node} summarizes the advantages and disadvantages of different node structures for cooperative perception. In a nutshell, V-PN is more ego-efficient (i.e., improving the perception capability from the standpoint of ego-vehicle.) while I-PN is more suitable for scalable cooperation. CP between homogeneous PNs, such as V2V or I2I, can mainly extend the perceptive range while CP between heterogeneous PNs, such as V2X, can achieve better FOV by complementing different sensor configurations.

\begin{table*}[!ht]
\centering
\caption{Performance matrix for different sensors utilized for infrastructure-based perception (rating range from 1 to 3 stars).}
\label{tab: sensors}
\resizebox{\textwidth}{!}{%
\begin{tabular}{l|l|l|l|l|l|l}
\toprule
\multicolumn{1}{c|}{Capabilities}             & \multicolumn{1}{c|}{Camera} & \multicolumn{1}{c|}{LiDAR} & \multicolumn{1}{c|}{RADAR} & \multicolumn{1}{c|}{Thermal} & \multicolumn{1}{c|}{Fisheye} & \multicolumn{1}{c}{Loop} \\ \midrule
Privacy-safe data                             & $\bigstar$                     & $\bigstar\bigstar\bigstar$          & $\bigstar\bigstar\bigstar$          & $\bigstar\bigstar$                 & $\bigstar$                      & $\bigstar\bigstar\bigstar$         \\
Accurately detects and classifies objects     & $\bigstar\bigstar$                & $\bigstar\bigstar\bigstar$          & $\bigstar$                    & $\bigstar\bigstar$                 & $\bigstar\bigstar$                 & $\bigstar$              \\
Accurately measures object speed and position & $\bigstar\bigstar$                & $\bigstar\bigstar\bigstar$          & $\bigstar\bigstar\bigstar$          & $\bigstar\bigstar$                 & $\bigstar\bigstar$                      & $\bigstar\bigstar$         \\
Extensive FOV  & $\bigstar$                     & $\bigstar\bigstar\bigstar$          & $\bigstar\bigstar$               & $\bigstar$                      & $\bigstar\bigstar\bigstar$            & $\bigstar$                   \\
Reliability across changes in lighting, sun, temperature     & $\bigstar\bigstar$                & $\bigstar\bigstar\bigstar$          & $\bigstar\bigstar\bigstar$          & $\bigstar\bigstar$                 & $\bigstar\bigstar$                 & $\bigstar\bigstar\bigstar$         \\
Ability to read signs and differentiate color & $\bigstar\bigstar\bigstar$           & $\bigstar\bigstar$               & $\bigstar$                    & $\bigstar$                      & $\bigstar\bigstar\bigstar$            & $\bigstar$                   \\ 
Cost for deployment and maintenance & $\bigstar\bigstar\bigstar$           & $\bigstar$               & $\bigstar\bigstar$                    & $\bigstar\bigstar$                      & $\bigstar\bigstar$            & $\bigstar$                   \\ 

\bottomrule
\end{tabular}%
}
\end{table*}

\section{Sensors Modality}
\label{sec: sensors}
For the CP system, sensors are the most fundamental modules due to their roles in raw data collection. Thus, this section overviews typical types of sensors that are utilized in transportation systems from different perspectives.
\subsection{Configuration and Performance}

For sensors equipped on current ADS, the most popular ones are cameras, LiDAR, and radar. Onboard radar has been deployed on vehicles to mainly achieve ADAS functionalities for many years~\cite{ziebinski2016survey}, such as Adaptive Cruise Control (ACC), Collision Avoidance, etc~\cite{tokoro2003electronically}.
For the onboard cameras, different ADS may take different configurations including single-camera ADS and multi-camera ADS. Single-camera ADS, such as the ADS developed by \textit{Comma.ai}~\cite{santana2016learning}, only deploys a camera-based perception system at the middle-top of the windshield. Multi-camera ADS, such as \textit{Waymo ADS}~\cite{sun2020scalability}, utilizes multiple cameras installed around the top-surrounding positions of the vehicle. In most common cases, LiDAR sensors, due to their capability of panoramic FOV, are mainly configured on the top of the vehicle. In some ADS, e.g., \textit{Waymo ADS}, auxiliary LiDAR systems are installed for complementing the blind zone of the top LiDAR~\cite{waymoADS}.

Regarding the installation of roadside sensors, typical locations may include signal arms and street lamp posts, with some minimum height requirements to avoid tampering. As a result, roadside sensors can have a much higher position (compared to onboard sensors) to minimize the occlusion effect due to dense traffic. The specific installation position may vary based on different roadside sensors. For example, the roadside LiDAR sensors are mainly installed at the height of $3-6m$ (but no more than $10m$), while fisheye cameras prefer a higher installation~\cite{bai2022cmm,bai2022cyber, zhao2019detection}. 

To form a comprehensive view of the general performance of different sensors used for perception in transportation systems, Table~\ref{tab: sensors} provides a summary of those that are widely utilized in ADS, traffic surveillance, and other transportation systems. Each of these sensors has its own capabilities and strengths in different use cases. 
\begin{itemize}
    \item Camera: High-resolution. Not great for 3D position and speed measurements, especially in dense traffic. 
    \item LiDAR: High-accuracy 3D perception with resilience to environmental changes. Not great with its relatively high price and data sparsity.
    \item RADAR: Measuring speed, unlocking applications like stop bar \& dilemma zone detection. Not great for distinguishing objects. 
    \item Thermal Camera: Getting thermal information, which provides resilience to lighting changes.
    \item Fisheye Camera: 360-degree full field-of-view (FOV) for detection. Requires a high-accurate calibration matrix to account for distortion.
    \item Loops: Measuring traffic counts and speed. Costly to install and maintain due to intrusiveness. 
\end{itemize}

\begin{figure}[!ht]
    \centering
    \includegraphics[width=0.5\textwidth]{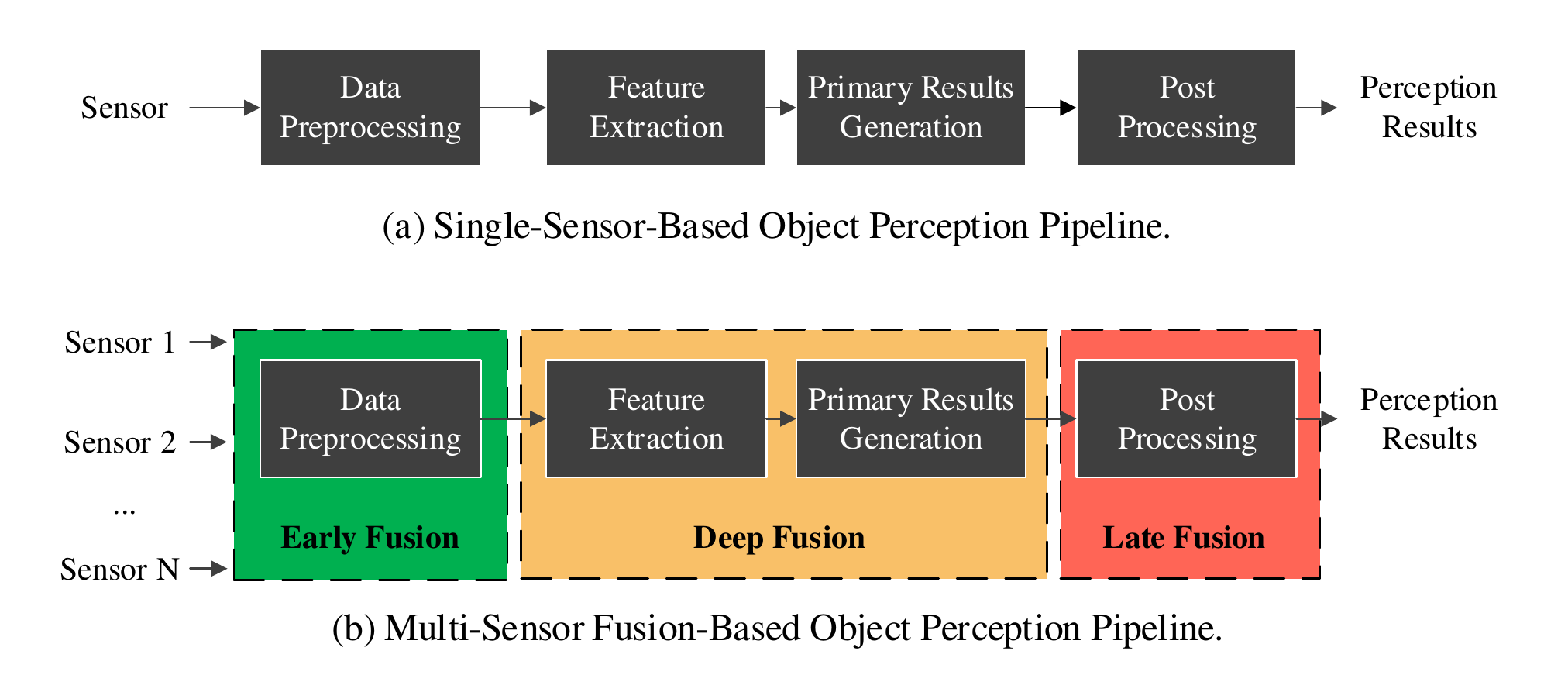}
    \caption{Systematic diagram of operational pipeline for: (a) single-sensor-based perception model; and (b) multi-sensor-based perception model.}
    \label{fig: pipelines}
\end{figure}

In terms of the number of sensors applied, a systematic operational pipeline of object perception can be divided into two main categories, i.e., single-sensor-based and multi-sensor-based, as shown in Fig.~\ref{fig: pipelines}.
\subsection{Single-Sensor Perception}
Single-sensor-based object perception systems have been widely developed and applied in the real-world transportation system whose main pipeline is demonstrated in Fig.~\ref{fig: pipelines}~(a). Data collected from the sensor is first \textit{preprocessed} to reduce noise, filter unrelated data, and properly reformat for downstream modules. Then, \textit{feature extraction} is applied to calculate predefined features by mathematical models (if based on traditional methods) or to generate hidden features by neural networks (if based on deep learning). Detection and tracking results are generated by the \textit{perception} module and are fed into the \textit{post-processing} module to further clean the perception outputs (e.g., filtering overlapped bounding boxes and predictions with scores under the threshold).

\label{sec: General Perception Methodologies}
In this section, we briefly cover major milestones of single-sensor perception chronologically from two perspectives -- the traditional approach and the deep-learning approach -- for cameras and LiDARs, respectively. 
\subsubsection{Camera}
Approximately twenty years ago, Viola and Jones~\cite{viola2004robust} proposed a method for real-time detection of human faces without any constraints. This algorithm outperformed any of other contemporary algorithms in terms of real-time performance, without compromising detection accuracy. In 2005, Dalal and Triggs~\cite{dalal2005histograms} proposed the Histogram of Oriented Gradients (HOG) feature descriptor which provided significant improvement of the scale-invariant feature transform~\cite{ lowe2004distinctive} and shapes context~\cite{belongie2002shape}. The HOG detector has been regarded as the cornerstone for many subsequent object detectors and implemented in various real-world applications~\cite{felzenszwalb2008discriminatively, felzenszwalb2010cascade}. Deformable Part-based Model (DPM) proposed by Felzenszwalb~et~al. ~\cite{felzenszwalb2008discriminatively} consecutively won the Pascal Visual Object Classes (VOC)-07, -08, and -09 detection challenges~\cite{everingham2010pascal}. Due to their dominant performance, DPM and its variants~\cite{felzenszwalb2010cascade} are widely regarded as the pinnacle of traditional object detection methods~\cite{zou2019object}.


Benefiting from the increased computational power, convolutional neural networks (CNNs)~\cite{krizhevsky2012imagenet} started to be widely used in 2012. Two years later, Girshick~et~al. proposed the Regions with CNN features (R-CNN) for object detection and completely unfolded the advantage of deep learning~\cite{girshick2015region}. In the same year, Spatial Pyramid Pooling Networks (SPPNet) proposed by He~et~al. was able to generate feature representation regardless of the image size, and run 20 times faster than R-CNN without compromising accuracy~\cite{he2015spatial}. In 2015, multiple renowned detectors were proposed by researchers: 1) Fast R-CNN~\cite{girshick2015fast} -- over 200 times faster than R-CNN -- proposed by Girshick; 2) Faster R-CNN~\cite{ren2015faster} -- the first end-to-end, and the first near-realtime deep learning detector -- proposed by Ren~et~al.; 3) You Only Look Once (YOLO)~\cite{redmon2016you} -- the first one-stage detector in the deep learning era with extremely fast speed (45 - 155 fps) -- proposed by Joseph~et~al.; and 4) Single Shot MultiBox Detector (SSD)~\cite{liu2016ssd} -- the second one-stage detector but with significantly improved accuracy -- proposed by Liu~et~al. In 2017, Lin~et~al. proposed Feature Pyramid Networks (FPN)~\cite{lin2017feature} based on Faster R-CNN, which achieved the SOTA object detection performance and has become a fundamental building block for various object perception models. In recent years, \textit{Transformers}~\cite{vaswani2017attention} embedded with the mechanism of attention have been leading the trend in the majority of object perception tasks, such as the Vision Transformer (ViT) proposed by Dosovitskiy~et~al.~\cite{dosovitskiy2020vit}, and Swin-Transformer proposed by Liu et~al.~\cite{liu2021swin}.

\subsubsection{LiDAR}
Before 2015, one of the most popular methodologies for solving PCD from LiDAR sensors is the bottom-up pipeline based on traditional methods, such as ``\textit{Clustering~\cite{ahmed2020density}$\rightarrow$Classification~\cite{zhang2016multilevel}$\rightarrow$Tracking~\cite{welch1995introduction}}''. Due to its explanation, interpolation, and free from data labeling, traditional bottom-up methodologies are still popular in current infrastructure-based LiDAR perception tasks~\cite{zhang2019vehicle,zhao2019detection,zhang2019automatic}. 

With the great success achieved by CNNs in image-based object perception, PCD quickly became the upcoming target for CNNs. However, PCD has a totally different data format compared with RGB images, which brings lots of challenges for applying existing CNN technologies to 2D vision tasks. Point-wise manipulation is considered straightforward for extracting features from PCD for object detection. \textit{Point RCNN}~\cite{shi2019pointrcnn} was proposed by Shi~et~al. to aggregate the point features via a \textit{PointNet++}~\cite{qi2017pointnet++} encoder. Endowed with the natural fit, point-based methods provide dominant performance in detection accuracy, however, under the sacrifice of computational efficiencies, such as \textit{PV-RCNN}~\cite{shi2020pv} (Shi~et~al. 2020).

Since PCD is in 3-dimension and sparse data, Wang~et~al.~\cite{wang2015voting} creatively cut the whole 3D point cloud into 3D voxels grids. Then a feature vector was designed to represent each voxel, which was fed into a linear SVM~\cite{suthaharan2016support} for classification results. A specific voting scheme was designed and mathematically proved to be able to act as sparse convolution~\cite{liu2015sparse} and the method was named \textit{Vote3D} in 2015. Two years later, Engelcke~et~al.~\cite{vote3ddeep} proposed the \textit{Vote3Deep}, which improved Vote3D by involving sparse convolution directly into the voting scheme. Furthermore, Rectified Linear Unit (ReLU)~\cite{nair2010rectified} and $L_{1}$ regularisation~\cite{vote3ddeep} (or named \textit{L1 Norm}) were involved to boost the learning process based on large sparse data, like PCD. In 2018, VoxelNet~\cite{zhou2018voxelnet} was proposed by Zhou~et~al., which introduced a learnable voxel encoder to generate hidden features for voxels. Specifically, 3D convolution was applied as a 3D backbone for 3D voxel feature extraction, and 2D CNN-based Region Proposal Network (RPN)~\cite{ren2016faster} was designed as a 2D backbone. This voxelization mechanism has been widely used in the following work, such as \textit{SECOND}~\cite{yan2018second} (Yan~et~al. 2018), \textit{PointPillar}~\cite{lang2019pointpillars} (Lang~et~al. 2019), \textit{Voxel RCNN}~\cite{deng2021voxel} (Deng~et~al. 2021), etc. 

Starting in 2018, projecting PCD into a 2D BEV feature map has quickly become a popular methodology. Inspired by YOLO, Simony~et~al.~\cite{simony2018complex} proposed \textit{ComplexYolo} which projected PCD into three manually defined feature channels, and then the BEV feature map was fed into a 2D backbone for generating detection results. Since the BEV scheme provides a straightforward way for solving 3D data in 2D manners, lots of BEV-based methods have emerged such as \textit{PIXOR}~\cite{yang2018pixor} (Yang~et~al. 2018), \textit{SCANet}~\cite{lu2019scanet} (Lu~et~al. 2019), \textit{BEVFusion}~\cite{liu2022bevfusion} (Liu~et~al. 2022), etc.

\subsection{Multi-Sensor Perception}
\label{sec: multi sensor}
Owing to the complementary of different sensors, multi-sensor-based perception systems have the potential to achieve better object detection and tracking performance via sensor fusion when compared with single-sensor-based perception systems. In this section, three popular multi-sensor perception schemes based on high-resolution sensors are discussed in this paper, i.e., \textit{Camera+Camera}, \textit{Camera+LiDAR}, and \textit{LiDAR+LiDAR}. 

\subsubsection{Cam + Cam}
The multi-camera system has been developed for decades and lots of applications have been designed and implemented in our current transportation systems~\cite{OlagokeCamSurvey}, such as object detection and object tracking.

For object detection, before the surge of CNN, the extraction and fusion of object-level features is a major challenge for traditional methods due to the high-dimensional complexity of RGB data. Merino~et~al.~\cite{merino2006cooperative} proposed a multi-UAV CP system based on heterogeneous sensor systems including infrared and visual cameras, fire sensors,
and others. A set of functions were designed for object detection including image segmentation, and stabilization of image sequences.
By coordinating the processed results from spatially separated sensors, the targeting object can be detected and localized based on a geo-referencing process. 

With the tremendous power of CNN to extract hidden features, object detection based on multi-camera systems quickly attracts lots of attention from researchers. For spatial alignment for the multi-node cameras, Arnold~et~al.~\cite{arnold2020cooperative} chose to project camera data from RGB images to pseudo-PCD. Owing to the 3D attribute of PCD, this pseudo-PCD could be easily aligned and merged into a unified coordinate system. Then a deep learning-based object detector was applied for generating perception results.

Object tracking has been widely developed in multi-camera systems for several decades to enable traffic surveillance and thus to analyze the traffic scenarios for further traffic optimization~\cite{wang2013intelligent}. The most typical way of multi-camera tracking is to calibrate the multi-camera systems to make all views stitched together in a unified coordinate system~\cite{eshel2008homography}. Meanwhile, consecutively tracking multi-objects under occluded conditions is one of the main strengths of a multi-camera tracking system which can provide sequences of images from different viewpoints. Specifically, based on the unified coordinate system gained from calibration, the Kalman Filter~\cite{mikic1998video}, the particle filter~\cite{kim2006multi}, etc., have been widely applied in multi-video object tracking systems.

The tracking schemes mentioned above generally require joint FOV for computing association across cameras. For the disjoint camera system, appearance cues are designed for capturing the common features between multiple views by integrating spatial-temporal information~\cite{song2008robust}. To overcome the dynamically changed spatial-temporal information in vision information, e.g., lighting condition and traffic speed, the tracking model should also be able to update its model adaptively. Thus, Expectation-Maximization (EM) framework~\cite{huang1997object}, unsupervised learning network~\cite{chen2008adaptive}, etc., have been implemented to dynamically update the model.


\begin{table*}[!ht]
  \centering
  \caption{Summary of Different Sensor Modalities for Cooperative Perception.}
  \resizebox{\textwidth}{!}{%
    \begin{tabular}{c|c|p{25em}|p{13em}|c}
    \toprule
    \multicolumn{1}{c|}{Structure} & 
    \multicolumn{1}{c|}{Modality} &
    \multicolumn{1}{c|}{Pros. and Cons.}  & 
    \multicolumn{1}{c|}{Highlighted Features} & 
    \multicolumn{1}{c}{Author} \\
    \midrule
    \multicolumn{1}{c|}{\multirow{4}[8]{*}{Single-Sensor}} & \multicolumn{1}{c|}{\multirow{2}[4]{*}{Camera}} & Pros: abundant vision data with cost-effective system. & \multicolumn{1}{c|}{\multirow{2}[4]{13em}{Using shifted window multi-head attention}} & \multicolumn{1}{c}{\multirow{2}[4]{*}{Liu~et~al.~\cite{liu2021swin}}} \\
\cmidrule{3-3}      &   & Cons: difficult to provide high-fidelity 3D information and significantly impact by lighting condition. &   &  \\
\cmidrule{2-5}      & \multicolumn{1}{c|}{\multirow{2}[4]{*}{Lidar}} & Pros: capable to provide high-fidelity 3D information with panoramic FOV. & \multicolumn{1}{c|}{\multirow{2}[4]{13em}{Encoding point cloud into voxelized pillars}} & \multicolumn{1}{c}{\multirow{2}[4]{*}{A. Lang,~et~al.~\cite{lang2019pointpillars}}} \\
\cmidrule{3-3}      &   & Cons: sparse data without vision information. &   &  \\
    \midrule
    \multicolumn{1}{c|}{\multirow{6}[12]{*}{Multi-Sensor}} & \multicolumn{1}{c|}{\multirow{2}[4]{*}{Cam. + Cam.}} & Pros: expand the FOV and perception area. & \multicolumn{1}{c|}{\multirow{2}[4]{13em}{Projecting RGB camera data into pseudo-LiDAR point cloud}} & \multicolumn{1}{c}{\multirow{2}[4]{*}{E. Arnold~et~al.~\cite{arnold2020cooperative}}} \\
\cmidrule{3-3}      &   & Cons: difficult to provide high-fidelity 3D information and significantly impact by lighting condition. &   &  \\
\cmidrule{2-5}      & \multicolumn{1}{c|}{\multirow{2}[4]{*}{Lidar + Lidar}} & Pros: expand the FOV and increase the density of the point cloud. & \multicolumn{1}{c|}{\multirow{2}[4]{13em}{Considering heterogeneous perception nodes, e.g., vehicle and infra.}} & \multicolumn{1}{c}{\multirow{2}[4]{*}{Bai~et~al.~\cite{bai2022pillargrid} }} \\
\cmidrule{3-3}      &   & Cons: sparse data without vision information. &   &  \\
\cmidrule{2-5}      & \multicolumn{1}{c|}{\multirow{2}[4]{*}{Cam. + Lidar}} & Pros: taking advantage of both camera and Lidar. & \multicolumn{1}{c|}{\multirow{2}[4]{13em}{Capturing BEV features from both sensors via CNN.}} & \multicolumn{1}{c}{\multirow{2}[4]{*}{Liu~et~al.~\cite{liu2022bevfusion}}} \\
\cmidrule{3-3}      &   & Cons: totally different information modality, thus difficult to fuse the data effectively. &   &  \\
    \bottomrule
    \end{tabular}}%
  \label{tab: sensor perception}%
\end{table*}%

\subsubsection{Cam + Lidar}
As different sensor modalities, camera and LiDAR seem to be a naturally complementary couple for perception. For instance, the camera is good at perceiving the vision information but lacking 3D distance data, while the LiDAR excels at collecting 3D information but lacking vision data. 

One typical way for the fusion of multi-modal sensor data is using CNN to extract hidden features in parallel and then combine them on the corresponding scale level. Zhu~et~al. proposed \textit{Multi-Sensor Multi-Level Enhanced YOLO} (\textit{MME-YOLO}) for vehicle detection in traffic surveillance~\cite{zhu2021mme}. MME-YOLO consists of two tightly coupled structures: 1) The enhanced inference head is empowered by attention-guided feature selection blocks and anchor-based/anchor-free ensemble head in terms of better generalization abilities in real-world scenarios; 2) The LiDAR-Image composite module is based on CBNet~\cite{liu2020cbnet} to cascade the multi-level feature maps from the LiDAR subnet to the image subnet, which strengthens the generalization of the detector in complex scenarios. MME-YOLO can achieve better performance for vehicle detection compared with YOLOv3~\cite{farhadi2018yolov3} for roadside sensor data.
 
Since camera and LiDAR have different poses and FOV, creating an intermediate feature level to unify LiDAR and image data before sending it to the feature-extraction backbone becomes a promising way for multi-modal sensor fusion. A popular way is to project camera information into LiDAR data to endow PCD with vision information. \textit{PointPainting}~\cite{vora2020pointpainting}, a point-level feature fusion method, decorates the PCD with semantic segmentation results from vision data. The point cloud data decorated with vision information are then fed into detectors, e.g., \textit{PointPillar}~\cite{lang2019pointpillars} for generating object detection results. Recently, Liu et~al.~\cite{liu2022bevfusion} proposed a novel framework, named \textit{BEVFusion}, to project both RGB and PCD information into a BEV feature map for fusion. Specifically, two dedicated encoders were designed to extract RGB and PCD inputs into the BEV feature map. Then, multi-modal feature fusion was conducted based on the spatial correspondence of BEV feature maps. The performance of \textit{BEVFusion} is the current SOTA for 3D object detection.

\subsubsection{Lidar + Lidar}
Although one single LiDAR can provide panoramic FOV around the ego-vehicle, physical occlusion may easily block the perceptive range and cause the ego-vehicle to lose some crucial perception information which significantly affects its decision-making or control process. On the other hand, a spatially separated LiDAR perception system can expand the perceptive range for intelligent vehicles or smart infrastructure.

One of the straightforward inspirations of the multi-LiDAR perception system is sharing the raw PCD via V2V communication~\cite{chen2019cooper}. However, limited wireless communication bandwidth may significantly limit real-time performance. Feature data generated from CNN requires much less bandwidth and is more robust to sensor noises, thus becoming a popular solution to multi-LiDAR fusion~\cite{F-cooper,wang2020v2vnet}. Marvasti~et~al.~\cite{marvasti2020cooperative} used two sharing-parameter CNNs to extract the feature map for PCD retrieved from two-vehicle nodes. Feature maps were then aligned based on the relative position and fused by element-wise summation. By applying an attention mechanism, Xu~et~al.~\cite{xu2021opv2v} proposed a V2V-based cooperative object detection method. A similar CNN process~\cite{lang2019pointpillars} was designed for extracting feature maps for V2V sharing. Furthermore, self-attention was involved in data aggregation based on spatial location in the feature map.

Recently, researchers started focusing on cooperation between V-PN and I-PN based on the multi-LiDAR system. For handling the data heterogeneity from roadside and onboard PCD, Bai~et~al.~\cite{bai2022pillargrid} proposed a decoupled multi-stream CNN framework for generating feature maps accordingly. Relative position information was applied to PCD alignment and the shared feature maps were then fused based on grid-wise \textit{maxout} operation. Additionally, Xu~et~al.~\cite{xu2022v2x} proposed a ViT-based CP method for heterogeneous PNs. Feature maps were extracted using sharing-parameter CNNs and V2X communications. For dealing with heterogeneity, specific graph transformer structures were designed for data extraction.

\begin{table*}[!ht]
  \centering
  \caption{Summary of Different Fusion Schemes for Cooperative Perception.}
  \resizebox{\textwidth}{!}{%
    \begin{tabular}{c|c|p{25em}|c|c}
    \toprule
    \multicolumn{1}{c|}{Fusion Scheme} & 
    \multicolumn{1}{c|}{Methodology} & 
    \multicolumn{1}{c|}{Pros. and Cons.} & 
    \multicolumn{1}{c|}{Highlighted Features} & 
    \multicolumn{1}{c}{Author} \\
    \midrule
    \multicolumn{1}{c|}{\multirow{2}[4]{*}{Early Fusion}} & \multicolumn{1}{c|}{\multirow{2}[4]{*}{Deep Learning}} & Pros: Raw data is shared and gathered to form a holistic view.  & \multicolumn{1}{c|}{\multirow{2}[4]{13em}{Raw point cloud data is compressed to fit the limited bandwidth.}} & \multicolumn{1}{c}{\multirow{2}[4]{*}{Chen~et~al.~\cite{chen2019cooper}}} \\
\cmidrule{3-3}      &   & Cons: Low tolerance to the noise and delay of the transmitted data; potentially constrained by the communication bandwidth. &   &  \\
    \midrule
    \multicolumn{1}{c|}{\multirow{2}[4]{*}{Deep Fusion}} & \multicolumn{1}{c|}{\multirow{2}[4]{*}{Deep Learning}} & Pros: High tolerance to the noise, delay, and difference between different nodes and sensor models. & \multicolumn{1}{c|}{\multirow{2}[4]{13em}{Deep neural features are extracted and fused based on spatial correspondence.}} & \multicolumn{1}{c}{\multirow{2}[4]{*}{Bai~et~al.~\cite{bai2022pillargrid}}} \\
\cmidrule{3-3}      &   & Cons: Require training data and hard to find a systematic way for model design. &   &  \\
    \midrule
    \multicolumn{1}{c|}{\multirow{2}[4]{*}{Late Fusion}} & \multicolumn{1}{c|}{\multirow{2}[4]{*}{Traditional}} & Pros: Easy to design and deploy in real-world system. & \multicolumn{1}{c|}{\multirow{2}[4]{13em}{A late-fusion is proposed based on joint re-scoring and non-maximum suppression.}} & \multicolumn{1}{c}{\multirow{2}[4]{*}{Zhang~et~al.~\cite{zhang2021distributed} }} \\
\cmidrule{3-3}      &   & Cons: Significantly limited by the wrong perception results or the difference between sources. &   &  \\
    \bottomrule
    \end{tabular}}%
  \label{tab: fusion}%
\end{table*}%

\subsection{Summary}
Table~\ref{tab: sensor perception} summarizes the advantages and disadvantages of different sensor modalities in the CP system. Different high-resolution sensors have different strengths. The camera is good at capturing vision information while LiDAR is excellent for collecting 3D information. Simultaneously taking advantage of these sensors in a complementary scheme is regarded as a promising solution to improving the perception accuracy of surveillance systems. 

\section{Fusion Scheme}
\label{fusion}
In terms of the stage of sensor fusion, a multi-sensor perception system can be divided into three classes: 1) \textit{Early Fusion} -- to fuse raw data at the preprocessing stage; 2) \textit{Deep Fusion} -- to fuse features at the feature extraction stage; and 3) \textit{Late Fusion} -- to fuse perception results at the post-processing stage. Different fusion schemes both have advantages and disadvantages in terms of different perspectives. For instance, Early Fusion and Deep Fusion have higher accuracy but need more computational power and complex model design. Conversely, Late Fusion can achieve better real-time performance but may sacrifice accuracy. It depends on the specific demands under different traffic scenarios to determine the best deployment of fusion schemes. This section aims to give a brief landscape of how fusion schemes are considered and applied in relevant CP research. Also, we will focus more on work that has not been introduced in previous sections.

\subsection{Early Fusion}
It is intuitive to share the raw sensor data with other PNs for expanding the perceptive range and improving detection accuracy. Following this strategy, the raw sensor data from multiple PNs are projected into a unified coordinate system for further processing~\cite{eshel2008homography}. However, since the basic idea of early fusion is only the expansion of raw data range or density, it is inevitably sensitive to the quality of sensor data, such as sensor calibration issues and data unsynchronization~\cite{wang2013intelligent}. Thus, early fusion can potentially provide the ideal performance only under several restricted assumptions, such as high-accurate sensor calibration and multi-source synchronization, which requires lots of effort in real-world implementations. 

On the other hand, early fusion requires large communication bandwidth to transmit a high volume of raw data. It is suitable for transmitting camera data with limited image resolution, but it may not be feasible to share real-time LiDAR data within a certain time delay (A 64-beam Velodyne LiDAR with 10Hz may generate about 20MB of data per second~\cite{Geiger2012CVPR}). For V2V early fusion, it is true that communicating raw sensor data  with one ego-vehicle is not an impossible solution~\cite{chen2019cooper}, but it is definitely not feasible for large-scale V2V cooperative perception under current communication capability.

\subsection{Late Fusion}
Standing in the opposite direction compared with early fusion, late fusion chooses another natural cooperative paradigm for perception -- generating perception results independently and then fusion them together. Different from early fusion, although late fusion also needs a relative position for fusing these perception results, its tolerance to calibration errors and unsynchronization issues is much higher than early fusion. One of the main reasons is that object-level fusion can be determined based on spatial and temporal constraints. For instance, Rauch et~al.~\cite{Car2X6232130} applied EKF to jointly align the shared bounding box proposals based on spatiotemporal constraints. Additionally, Non-Maximum Suppression (NMS)~\cite{neubeck2006efficient} and other machine-learning-based proposal refining methods are widely applied in late fusion methods for object perception~\cite{arnold2020cooperative}. Recently, due to the distributed attributes of late fusion, \textit{Federated Learning}~\cite{liu2020fedvision} also attracts 
increasing popularity in perception systems~\cite{zhang2021distributed}.

\begin{figure*}[!h]
    \centering
    \includegraphics[width=\textwidth]{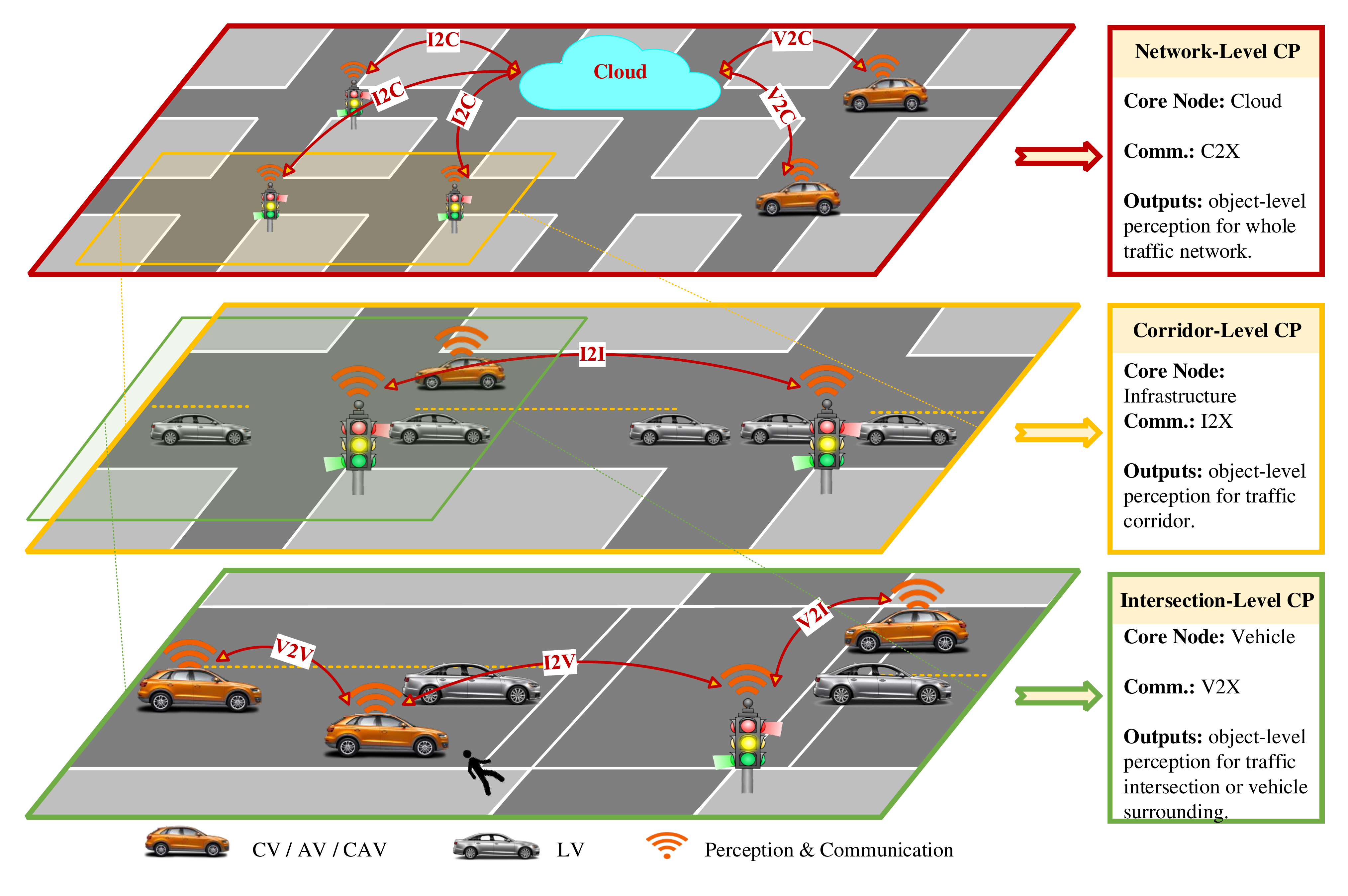}
    \caption{The schematic diagram of the HCP framework.}
    \label{fig: HCP}
\end{figure*}

\subsection{Deep Fusion}
The core ideology of deep fusion (also named \textit{Intermediate Fusion}) can be simply summarized as using deeply extracted features for fusion that happens at the intermediate stages of the perception pipeline. Deep fusion relies on hidden features mainly extracted from deep neural networks, which have higher robustness compared with raw sensor data used for early fusion. Xu~et~al.~\cite{xu2022v2x} assessed the robustness of model performance under different time delays and noises of metadata (the ego-vehicle location and heading). Different levels of errors were involved in the cooperative perception process. The evaluation results can be summarized as three points: 
\begin{itemize}
    \item With no error involved, early fusion and deep fusion can achieve similar performance which is better than late fusion;
    \item With the increase of errors, the performance of both early fusion and late fusion decreases drastically, but the performance degradation of all deep fusion methods~\cite{F-cooper, wang2020v2vnet,xu2021opv2v, xu2022v2x} is much less noticeable than early fusion and late fusion.
\end{itemize}
Additionally, feature-based fusion methods typically have only one detector for generating object perception results and thus there is no need for merging multiple proposals as required by late fusion~\cite{zhang2021distributed, arnold2020cooperative}. 

Although cooperative perception has been developed in multiple areas for several decades, deep-fusion-based cooperative perception is still in its infancy. Most of the deep fusion methods for CP were devised in the past few years, such as \textit{F-Cooper}~\cite{F-cooper} (2019), \textit{V2V Net}~\cite{wang2020v2vnet} (2020), \textit{OPV2V}~\cite{xu2021opv2v} (2021), \textit{PillarGrid}~\cite{bai2022pillargrid} and \textit{V2X-ViT}~\cite{xu2022v2x} (2022), etc. So far, most of the deep feature extraction is conducted by CNN, such as ~\cite{F-cooper, wang2020v2vnet, bai2022pillargrid}, because the CNN-based feature is highly related to the local spatial information. Very recently, some studies have applied transformers as the deep feature extractor~\cite{xu2021opv2v, xu2022v2x} due to their capability for panoramic feature extraction.

\subsection{Summary}
Table~\ref{tab: fusion} summarizes the advantages and disadvantages of different sensor fusion schemes for CP systems. Early fusion only needs the calibration for aligning multi-source data into a unified coordinate system but requires a large communication bandwidth for transmitting data. Late fusion mainly focuses on how to merge the proposals generated from multiple perception pipelines, which is straightforward but suffered from limited accuracy. Deep fusion is quickly becoming a transformable solution for CP due to its capabilities of low-communication requirements and high accuracy.

\section{Hierarchical Cooperative Perception Framework}
\label{sec: hcp}
Based on the overview of the aforementioned literature, Three major issues can be identified for CP system in the real world:
\begin{itemize}
    \item \textbf{Heterogeneity}: the CP system should take advantage of both intelligent vehicles and smart infrastructures to empower the comprehensiveness of perception.
    \item \textbf{Scalability}: the CP system needs to be able to extend to different scales of cooperation levels, such as intersection level, corridor level, and traffic network level. 
    \item \textbf{Dynamism}: the CP system needs to be able to dynamically cooperate with vehicle perception nodes, i.e, the I-PN should be capable of consecutively cooperating with a dynamically changed number of V-PNs.
\end{itemize}
    
To address the issues mentioned above, we propose a unified CP framework, called \textit{Hierarchical Cooperative Perception} (HCP) Framework, which is demonstrated in Fig.~\ref{fig: HCP}. HCP aims to assimilate different CP tasks under various scenarios into a general framework. The design of the HCP framework is based on 1) the system architecture for CP as shown in Fig.~\ref{fig: roadside perception structure}, 2) the taxonomy of CP as shown in Fig.~\ref{fig: taxonomy}, and 3) the analysis of reviewed literature. 

In this paper,  the HCP framework mainly focuses on the intersection scenarios and consists of three-level: 1) Intersection-Level CP, 2) Corridor-Level CP, and 3) Network-Level CP, which will be introduced from several perspectives including core node, communication types, and perception outputs, respectively. 

\subsubsection{Intersection-Level CP}
As shown in the bottom part of Fig.~\ref{fig: HCP}, intersection-level CP aims to perceive the object-level traffic condition around an intersection. V-PN and I-PN are designed as the core perception node at this level. For vehicles that are equipped with powerful onboard processors such as CAVs, features can be shared via V2V communication and processed onboard. The perception results from I-PN can act as auxiliary data to augment the CAV's perception results by late fusion. Most of the previous V2V CP work~\cite{wang2020v2vnet, xu2021opv2v, F-cooper} can be integrated into our HCP framework from this perspective.

Since the edge processor can be deployed at the I-PN for processing the roadside sensor data and the data received from intelligent vehicles via V2I communication. Vehicles are not necessarily required to be equipped with a powerful onboard processor for processing the whole perception pipeline. Lightweight computing units can be deployed for only extracting the feature. Deep features from multiple vehicles can be transmitted to the I-PN for deep fusion to generate perception results. The I-PN then broadcasts the perception results to vehicles within its own communication range. Recent V2I-based CP can be regarded as a specific version of the intersection-level CP~\cite{bai2022pillargrid, xu2022v2x}. Intersection-level CP is a crucial component for unlocking the current bottleneck (in terms of efficiency, safety, and sustainability) for cooperative driving automation in a mixed traffic environment~\cite{bai2022hybrid}.

\subsubsection{Corridor-Level CP}
As shown in the middle of Fig.~\ref{fig: HCP}, corridor-level CP aims to expand the perception based on the connectivity of multiple smart infrastructures. The core is the infrastructure node, i.e., I-PN. Currently, I2I communication (via cable or optical fiber) has a much higher capacity compared with wireless communication. For instance, optical fiber can achieve over $40GB/s$ communication speed with low latency and even commercial optical-fiber internet can achieve $1GB/s$~\cite{poletti2013towards}, which is enough for transmitting intersection-level data between intersections.

Empowered by high-speed communication, I2I-based CP is capable of applying all fusion schemes based on specific scenarios. Raw data sharing can be a typical style for I2I-based CP~\cite{arnold2020cooperative}. Meanwhile, by sharing feature-level data with corridor-level I-PNs, the CP system can generate object-level perception information with high perception accuracy to further assist road users or improve traffic management~\cite{bai2022cyber}.

\subsubsection{Network-Level CP}
As shown at the top of Fig.~\ref{fig: HCP}, network-level CP aims to perceive the object-level traffic condition for the whole traffic network. The cloud server is the core node to link all distributed intersections and CAVs that are out of the I-PN range. The cost-effective way for network-level CP is late fusion -- retrieving perception information from I-PNs and CAVs and then merging those results for distribution.

Furthermore, feature-level data can be also transmitted to the cloud server and a unified detector can be designed for generating the perception results.

\section{Datasets and Simulators}
In this section, we briefly introduce the tools that support the development of cooperative perception including datasets and simulators. We hope this section can give researchers a quick glance at the foundations that can possibly enable their relevant research.
\subsection{Datasets}
\subsubsection{General Object Perception}
Owing to prevailing needs in autonomous driving for surrounding perception, most real-world datasets for object detection and tracking are collected from onboard sensors. Several widely used datasets (both supporting Camera and LiDAR) for driving automation are briefly introduced as follows:
\begin{itemize}
    \item \textit{KITTI}: one of the most popular datasets, which consists of hours of traffic scenarios recorded with a variety of sensor modalities for mobile robotics and autonomous driving~\cite{Geiger2012CVPR}. 
    \item \textit{NuScenes}: the first dataset to carry the fully autonomous vehicle sensor suite: 6 cameras, 5 radars, and 1 LiDAR, all with a full 360-degree field of view~\cite{caesar2020nuscenes}.
    \item \textit{Waymo Open Dataset}: a large-scale, high-quality, diverse dataset that consists of 1150 scenes captured across a range of urban and suburban geographical terrains~\cite{sun2020scalability}.
\end{itemize}

\subsubsection{Infrastructure-based Perception}
Because roadside perception has great potential to promote the development of CDA, there are immediate demands for establishing a roadside sensor-based dataset for various infrastructure-based object perception tasks. In 2021, \textit{BAAI-VANJEE Roadside Dataset} was published by Deng~et~al. to support the Connected Automated Vehicle Highway technologies~\cite{yongqiang2021baai}. The BAAI-VANJEE Roadside Dataset consists of LiDAR data and RGB images collected by a roadside data-collection platform and contains 2500 frames of LiDAR data, and 5000 frames of RGB images which includes 12 classes of objects, 74K 3D object annotations, and 105K 2D object annotations.

\subsubsection{Cooperative Perception}
Although various kinds of real-world datasets have been collected for training models for different perception tasks. Before 2022, there is no available open-sourced cooperative perception dataset for real-world data. Thus, to overcome this issue, researchers mainly follow two ways of dataset acquisition. The most popular way is to build cooperative perception scenarios in a high-fidelity simulator and then collect multi-node multi-sensor data from the environment. Several datasets have been built based on \textit{CARLA}~\cite{dosovitskiy2017carla}. For instance, to enable V2V cooperative perception, \textit{OpenV2V} Dataset~\cite{xu2021opv2v} is collected by attaching LiDAR sensors to multiple vehicles in the CARLA simulator under different scenarios. 

Additionally, for heterogeneous perception nodes, the \textit{CARTI} dataset~\cite{bai2022pillargrid, CARTI2022} has been collected by deploying LiDAR and camera sensors to both vehicles and infrastructure in CARLA environments. Specifically, the raw data, sensor calibration, and ground truth label are designed following the same format as \textit{KITTI} dataset. Thus \textit{CARTI} dataset is readily integrated into the current deep learning codebase for quick development, such as \textit{MMDetction3D}~\cite{mmdet3d2020}. 

In 2022, \textit{DAIR-V2X}~\cite{yu2022dair}, the first real-world cooperative perception dataset comes to the stage, which is a large-scale, multi-node, multi-modality CP dataset. Specifically,  \textit{DAIR-V2X} contains $39k$ images, $39k$ PCD frames, and $10$ classes of ground truth labels with synchronized time stamps. Sensors are collected from both vehicle nodes and infrastructure nodes.

\subsection{Simulators}
In the context of cooperative perception using simulation, high-fidelity sensors are inevitably required due to the dependence on high-resolution sensor data~\cite{kiran2021deep}. Although traditional microscopic traffic simulators can also emulate the behavior at the object level, such as SUMO~\cite{behrisch2011sumo}, VISSIM~\cite{fellendorf2010microscopic}, AIMSUN~\cite{barcelo2005dynamic}, etc, they can not provide high-resolution sensor data with high fidelity. Thus, in recent years, several autonomous driving simulators have been developed to enable high-fidelity modeling of the surrounding environment and sensor capability by utilizing game engines, such as Unity \cite{juliani2018unity} and Unreal Engine \cite{karis2013real}. Specifically, several representative simulators are  CARLA~\cite{dosovitskiy2017carla}, SVL~\cite{rong2020lgsvl},  AirSim~\cite{shah2018airsim}, etc. 

\begin{itemize}
    \item CARLA is an open-source simulator for autonomous driving and supports flexible specifications of sensor suites and environmental conditions. In addition to open-source codes and protocols, CARLA provides open digital assets (e.g., urban layouts, buildings, and vehicles) that can be used in a friendly manner for researchers. 
    \item SVL is a high-fidelity simulator for driving automation, which provides end-to-end and full-stack simulation ready to be hooked up with several open-source autonomous driving stacks, such as Autoware \cite{kato2018autoware} and Apollo \cite{graf2008apollo}. 
    \item Besides high-fidelity sensors and environments, AirSim includes a physics engine that can operate at a high frequency for real-time hardware-in-the-loop (HIL) simulations with the support for popular protocols, such as MavLink \cite{koubaa2019micro}. 
\end{itemize}

These simulators are all open-source with detailed tutorials and can provide high-resolution and high-fidelity sensor data, such as cameras and LiDARs. These simulators can provide a highly customized and cost-effective way for collecting training datasets and traffic scenarios, and thus are widely applied in learning-based object perception tasks~\cite{ arnold2020cooperative, arnold2019survey}.

\section{Discussion}
\label{sec: dis}
Although cooperative perception is an emerging research area, it is playing an increasingly significant role in promoting the perception capabilities for CDA applications. Many studies have been conducted to lay the foundation and provide inspiration for future work. In this section, we present our insights concerning the current states, open problems, and future trends in cooperative perception for CDA applications.
\subsection{Current States and Open Challenges}
\subsubsection{Perception Singleton for Heterogeneity}
The most common perception agents in transportation are intelligent vehicles and smart infrastructure which can be regarded as heterogeneous perception singletons. Since roadside sensors have more flexible locations and pose for data acquisition, one typical way of cooperative perception is to transmit information from the infrastructure side to road users~\cite{balamuralidhar2021multeye, 9502706, zhang2020gc, bai2022cmm}. From the perspective of cooperative automated driving, V2V-based cooperative perception is also a promising solution to enabling the ego-vehicle with the capability of \textit{seeing through}~\cite{rockl2008v2v, Car2X6232130, Machine8569832, kim2014multivehicle}.

However, none of them can make an epochal revolution if they do not cooperate together in a deep manner, because the evolution of intelligent transportation systems is always highly coupled with the cooperation between vehicles and infrastructures~\cite{5959985}. Due to the heterogeneity of the perception singleton, only recently few studies have considered the cooperation between vehicle nodes and infrastructure nodes~\cite{bai2022pillargrid, xu2022v2x}. Thus, vehicle-infrastructure cooperation is one of the most significant opening tasks for cooperative perception.

\subsubsection{Sensor System for Fidelity}
To the most extent, the capability of the sensor system can regarded as the stepping stone of an intelligent transportation system. Since the perception data generated from sensor systems is the foundation of the downstream modules, such as prediction, decision-making, and actuation~\cite{bai2022cyber}. Thus for cooperative perception, cameras and LiDAR are widely applied to accessing high-fidelity sensing data. 

However, in most of the research, these two kinds of high fidelity sensors work separately -- a cooperative perception system only equipped with one kind of sensor -- such as multi-camera-based CP~\cite{zhu2021mme, arnold2020cooperative} and multi-LiDAR-based CP~\cite{wang2020v2vnet,bai2022pillargrid}. According to the analysis in Section~\ref{sec: multi sensor}, cameras and LiDAR are naturally complimentary to each other, and the camera-LiDAR-based perception method can also achieve the SOTA performance in general object detection~\cite{liu2022bevfusion}. Thus, developing multi-modality sensors for cooperative perception is an important way to improve the overall fidelity of the perception results. 

On the other hand, although the infrastructure plays a key role in cooperative perception, current roadside-sensor-based perception methods are, to most extent, applied directly from general perception methods, i.e., onboard sensor-based model. Comparing the methods reviewed in Section~\ref{sec: sensors} and Section~\ref{sec: Node}, there is an evident gap between general object perception and cooperative perception. For instance, the core methodologies of a large portion of the existing roadside LiDAR-based detection approaches are based on DBSCAN for clustering~\cite{Song9216093, 8484040, zhang2019automatic, zhao2019detection, zhang2020vehicle}, which has a performance gap compared with the SOTA methods~\cite{lang2019pointpillars, zhou2018voxelnet}. Since the sensing methodologies of roadside sensors are different from onboard perceptions, one of the major challenges is roadside data acquisition and annotation for promoting the deep learning-based research of infrastructure-based perception systems.

\subsubsection{Fusion Strategies for Generality}
As reviewed in Section~\ref{fusion}, different fusion schemes have their specific advantages and disadvantages. Early fusion-based studies mainly require high-speed communication to enable the transmission of raw data~\cite{chen2019cooper, arnold2020cooperative}. However, the reliance on raw data inevitably makes the perception model very sensitive, and small communication errors or synchronization issues can cause significant degradation in system performance~\cite{xu2022v2x}. Late fusion-based research has been widely applied to various kinds of cooperative perception tasks since decades ago~\cite{merino2006cooperative, arnold2020cooperative, zhang2021distributed}. Late fusion has less requirement for communication but its performance also suffers from the merging of the object proposals from multiple sources~\cite{arnold2020cooperative}. 

To solve the issues mentioned above, recent work has been focusing on transmitting and fusing feature-level data to gain better accuracy with higher robustness~\cite{bai2022pillargrid, xu2022v2x}. However, due to the deeply coupled feature and model complexity, large-scale extension is an inevitable challenge for deep fusion-based cooperative perception.


\subsection{Future Trends}
\subsubsection{Towards Heterogeneous Cooperation}
Physical occlusion is considered one of the unavoidable obstacles to single-node perception, and perceiving the environment from multiple nodes can mitigate such limitations. Given that transportation is a system of systems, vehicle-infrastructure cooperation is a promising solution to many existing traffic-related issues. More specifically, vehicle-infrastructure cooperative perception can leverage the capabilities of both vehicles (as mobile perception nodes with lightweight processing power) and infrastructure (as fixed nodes but with powerful processing/storage units) to achieve much better performance. Efficient and dynamic ways to cooperate the information from vehicles with infrastructures are the keys to unlocking a new era of perception for cooperative driving automation.

\subsubsection{Towards Multi-Modal Cooperation}
A multi-sensor-based perception system has the potential to improve perceived performance by taking advantage of complementary sensor data~\cite{zamanakos2021comprehensive} with appropriate fusion techniques. In the scope of camera and LiDAR sensors, the development of current multi-modal sensor fusion is mainly targeting general object perception by multiple sensors equipped on one single agent~\cite{liu2022bevfusion}. Specific multi-modal sensor fusion for multiple perception nodes is still a blank field, which is, however, an important way to improve the perception accuracy for the whole system.

\subsubsection{Towards Scalable Cooperation}
The concept of cooperative perception is never intended to be only applied to a small number of nodes, such as two vehicles~\cite{F-cooper} or one vehicle with one infrastructure~\cite{bai2022pillargrid}. Some cooperative perception methods are mainly designed for enhancing the ego-vehicle with the assistance of surrounding nodes by asking surrounding nodes to align their data based on the metadata from the ego-vehicle~\cite{xu2022v2x}, which may cause scalability issues when numerous ego-vehicles are involved. 

On the other hand, the computational power and perceptive range of perception nodes are not the same for vehicles and infrastructure. An infrastructure-based perception system is more flexible in terms of sensor equipment and capable of empowering high-computational edge processors, large data storage and wide communication bandwidth. Although the onboard device has made major strides in development, it could be extremely costly and energy inefficient to empower every single vehicle with a high-performance computation system for perception. Therefore, by only deploying lightweight onboard computation modules on the vehicle side, such as feature map extraction, it becomes much more cost-effective to 1) enable local deep-fusion-based cooperative perception~\cite{bai2022pillargrid} or 2) retrieve perception results from infrastructure-based high-performance nodes for a wider range of perceptions~\cite{bai2022cmm}.

Considering the issues for cooperative perception in real-world development, such as scalability, dynamic environment, and heterogeneous resources (such as computational power, storage space, and communication bandwidth), a hierarchical structure, including vehicle, infrastructure, and cloud, introduced in Section~\ref{sec: hcp} can be a promising solution. Thus, building a unified framework will be a systematic challenge and can lay a solid foundation for further research on cooperative perception.


\section{Conclusions}
\label{sec: conc}
This paper provides a comprehensive overview and proposes a hierarchical framework for cooperative perception. The architecture and taxonomy are presented to illustrate the fundamental components and core aspects of the cooperative perception system. Cooperative perception methods are then introduced with detailed literature reviews from three perspectives: node structure, sensor modality, and fusion scheme. The proposed hierarchical cooperative perception framework is analyzed from the various levels of intersection, corridor, and network respectively. Existing datasets and simulators for enabling cooperative perception are briefly reviewed to identify the gaps. Finally, this paper discusses current issues and future trends. To the best of our knowledge, this work is the first study to provide a unified framework for cooperative perception.

\section*{Acknowledgments}
This research was funded by Toyota Motor North America, InfoTech Labs. The contents of this paper reflect the views of the authors, who are responsible for the facts and the accuracy of the data presented herein. The contents do not necessarily reflect the official views of Toyota Motor North America.

\bibliographystyle{IEEEtran}  
\bibliography{references}

\begin{IEEEbiography}
    [{\includegraphics[width=1in,height=1.25in,clip,keepaspectratio]{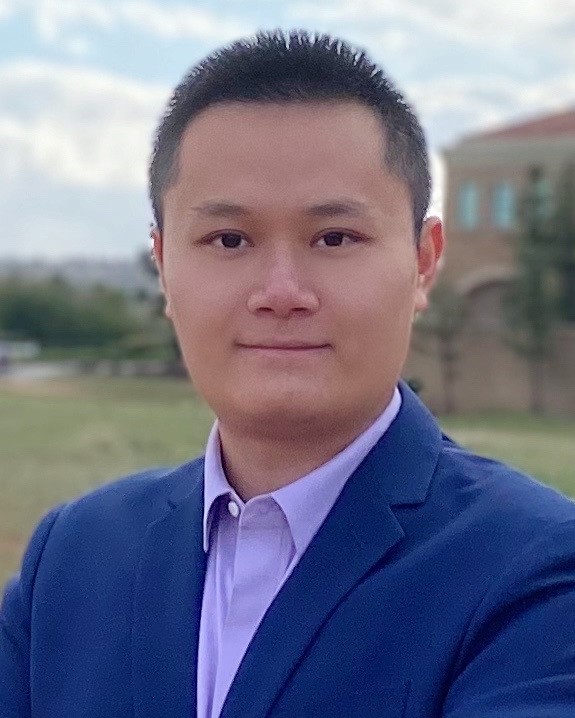}}]{Zhengwei Bai}
(Student Member, IEEE) received the B.E. and M.S. degrees from Beijing Jiaotong University, Beijing, China, in 2017 and 2020, respectively. He is currently a Ph.D. student in electrical and computer engineering at the University of California at Riverside. His research focuses on object detection and tracking,  cooperative perception, decision making, motion planning, and cooperative driving automation (CDA). He serves as a Review Editor in Urban Transportation Systems and Mobility.
\end{IEEEbiography}

\begin{IEEEbiography}
    [{\includegraphics[width=1in,height=1.25in,clip,keepaspectratio]{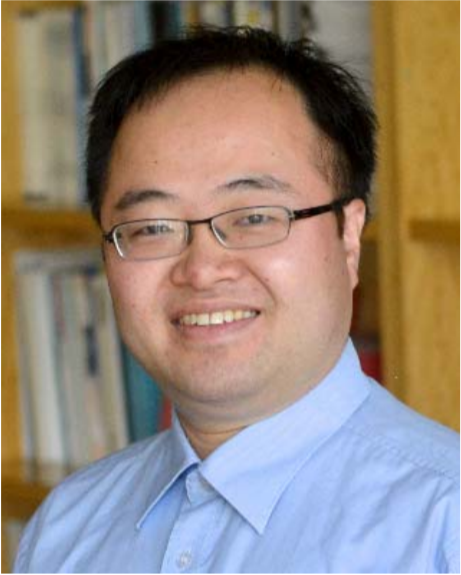}}]
    {Guoyuan Wu}
(Senior Member, IEEE) received his Ph.D. degree in mechanical engineering from the University of California, Berkeley in 2010. Currently, he holds an Associate Researcher and an Associate Adjunct Professor position at Bourns College of Engineering – Center for Environmental Research \& Technology (CE–CERT) and Department of Electrical \& Computer Engineering in the University of California at Riverside. development and evaluation of sustainable and intelligent transportation system (SITS) technologies, including connected and automated transportation systems (CATS), shared mobility, transportation electrification, optimization and control of vehicles, traffic simulation, and emissions measurement and modeling. Dr. Wu serves as Associate Editors for a few journals, including IEEE Transactions on Intelligent Transportation Systems, SAE International Journal of Connected and Automated Vehicles, and IEEE Open Journal of ITS. He is also a member of the Vehicle-Highway Automation Standing Committee (ACP30) of the Transportation Research Board (TRB), a board member of Chinese Institute of Engineers Southern California Chapter (CIE-SOCAL), and a member of Chinese Overseas Transportation Association (COTA). He is a recipient of Vincent Bendix Automotive Electronics Engineering Award.
\end{IEEEbiography}

\begin{IEEEbiography}
    [{\includegraphics[width=1in,height=1.25in,clip,keepaspectratio]{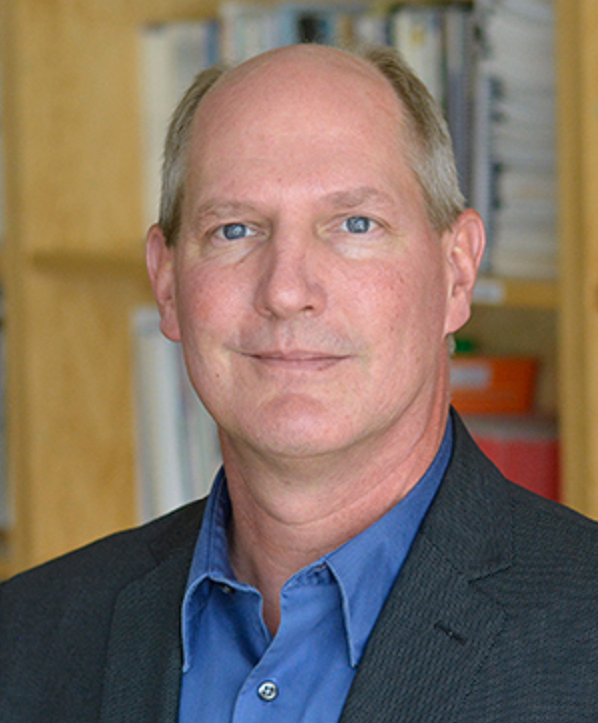}}]
    {Matthew J. Barth}
(Fellow, IEEE) received the M.S. and Ph.D degree in electrical and computer engineering from the University of California at Santa Barbara, in 1985 and 1990, respectively. He is currently the Yeager Families Professor with the College of Engineering, University of California at Riverside, USA. He is also serving as the Director for the Center for Environmental Research and Technology. His current research interests include ITS and the environment, transportation/emissions modeling, vehicle activity analysis, advanced navigation techniques, electric vehicle technology, and advanced sensing and control. Dr. Barth has been active in the IEEE Intelligent Transportation System Society for many years, serving as a Senior Editor for both the Transactions of ITS and the Transactions on Intelligent Vehicles. He served as the IEEE ITSS President for 2014 and 2015 and is currently the IEEE ITSS Vice President of Education.
\end{IEEEbiography}

\begin{IEEEbiography}
    [{\includegraphics[width=1in,height=1.25in,clip,keepaspectratio]{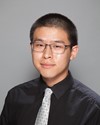}}]{Yongkang Liu}
received the Ph.D. and M.S. degrees in electrical engineering from the University of Texas at Dallas in 2021 and 2017, respectively. He is currently a Research Engineer at Toyota Motor North America, InfoTech Labs. His current research interests are focused on in-vehicle systems and advancements in intelligent vehicle technologies.  
\end{IEEEbiography}

\begin{IEEEbiography}
    [{\includegraphics[width=1in,height=1.25in,clip,keepaspectratio]{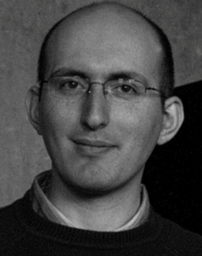}}]{Emrah Akin Sisbot}
(Member, IEEE) received the Ph.D. degree in robotics and artificial intelligence from Paul Sabatier University, Toulouse, France in 2008. He was a Postdoctoral Research Fellow at LAAS-CNRS, Toulouse, France, and at the University of Washington, Seattle. He is currently a Principal Engineer with Toyota Motor North America, InfoTech Labs, Mountain View, CA. His current research interests include real-time intelligent systems, robotics, and human-machine interaction.
\end{IEEEbiography}

\begin{IEEEbiography}
    [{\includegraphics[width=1in,height=1.25in,clip,keepaspectratio]{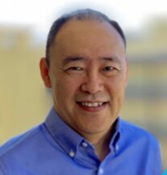}}]
{Kentaro Oguchi} received the M.S. degree in computer science from Nagoya University. He is currently a Director at Toyota Motor North America, InfoTech Labs. Oguchi’s team is responsible for creating intelligent connected vehicle architecture that takes advantage of novel AI technologies to provide real-time services to connected vehicles for smoother and efficient traffic, intelligent dynamic parking navigation and vehicle guidance to avoid risks from anomalous drivers. His team also creates technologies to form a vehicular cloud using Vehicle-to-Everything technologies. Prior, he worked as a senior researcher at Toyota Central R\&D Labs in Japan.
\end{IEEEbiography}

\begin{IEEEbiography}
    [{\includegraphics[width=1in,height=1.25in,clip,keepaspectratio]{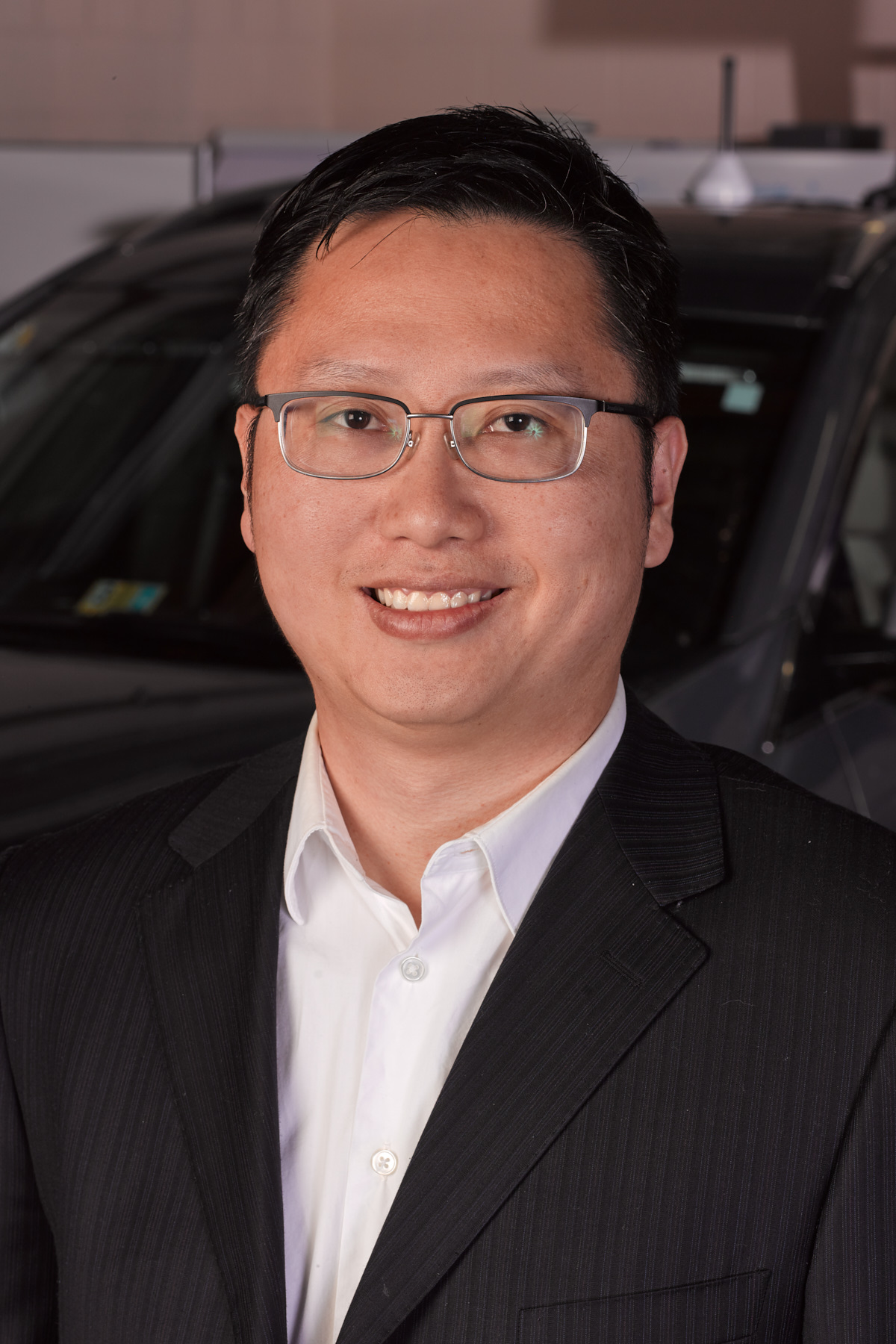}}]
{Zhitong Huang} is senior transportation research scientist and Analysis, Simulation, and Modeling program manager at Leidos. He has 17 years of research experience and conducted dozens of research projects in the field of transportation engineering. His main focus is on transportation simulation and modeling, connected and automated vehicle (CAV) systems, traffic operation and management, and digital twin, etc. 
\end{IEEEbiography}
\end{document}